%% file: paper.tex
\documentclass{article}

\usepackage{iclr2027_conference,times}
\input{math_commands.tex}

\usepackage{hyperref}
\usepackage{etoc}
\usepackage{url}
\usepackage{graphicx}
\usepackage{booktabs}
\usepackage{caption}
\usepackage[most]{tcolorbox}
\iclrfinalcopy
\newtcolorbox{swpmprompt}[1]{
  colback=cyan!3,colframe=cyan!45!black,
  title={#1},breakable,fontupper=\footnotesize,
  fonttitle=\small\bfseries,boxrule=0.5pt,arc=1.5mm,
  left=7pt,right=7pt,top=5pt,bottom=5pt}

\title{Shared Worlds, Private Minds:\\
Structured Memory for Long-Form Writing as World Creation}

\hypersetup{
  colorlinks=true,
  linkcolor=blue,
  citecolor=blue,
  urlcolor=blue,
  pdftitle={Shared Worlds, Private Minds: Structured Memory for Long-Form Writing as World Creation},
  pdfauthor={Qiuyu Tian, Xiaowen Gu, Hang Su, Jianghan Chao, Haojie Yin, Fan Guo, Xin Zhang, Jinjing Shen, Ewing Luo, Youyong Kong, Yingce Xia, Zequn Liu}
}

\author{%
\normalfont
\begin{tabular}{@{}c@{\quad}c@{\quad}c@{\quad}c@{}}
\textbf{Qiuyu Tian}$^{1,2}$ &
\textbf{Xiaowen Gu}$^{6,7}$ &
\textbf{Hang Su}$^{2,4}$ &
\textbf{Jianghan Chao}$^{2,5}$ \\
\textbf{Haojie Yin}$^{3}$ &
\textbf{Fan Guo}$^{8}$ &
\textbf{Xin Zhang}$^{8}$ &
\textbf{Jinjing Shen}$^{8}$ \\
\textbf{Ewing Luo}$^{8}$ &
\textbf{Youyong Kong}$^{1}$ &
\textbf{Yingce Xia}$^{2}$ &
\textbf{Zequn Liu}$^{2,*}$ \\[0.45em]
\multicolumn{4}{c}{%
\begin{tabular}{c}
\small $^{1}$Southeast University, Nanjing, China \\
\small $^{2}$Beijing Zhongguancun Academy, Beijing, China \\
\small $^{3}$Medical Physics, Duke University \\
\small $^{4}$East China Normal University \\
\small $^{5}$Gaoling School of Artificial Intelligence, Renmin University of China \\
\small $^{6}$School of Animation and Digital Arts, Communication University of China \\
\small $^{7}$Jiangsu Second Normal University \\
\small $^{8}$ZhuiWen Technology Co., Ltd., Beijing, China \\
\small $^{*}$Corresponding author.
\end{tabular}}
\end{tabular}%
}

\begin{document}

\maketitle

\begin{abstract}
LLM agents that write long-form fiction need an explicit memory of the evolving storyworld to keep new events consistent with established facts. Such memory must keep heterogeneous narrative information distinct, integrate story developments across granularities, and recover dependencies that a writing request leaves implicit. We present NarraWorld, a structured memory system for long-form writing that treats memory construction as world creation. From a shared evidence-grounded graph, NarraWorld derives four connected views: world facts, per-character beliefs, open developments, and hypothetical branches (possible-world continuations). Hierarchical aggregation with atomic closure consolidates events into scenes, plotlines, and plots, keeping each higher-level node traceable to its constituent source spans. For retrieval, planned reconstruction infers a query's dependencies from the current narrative situation and a preview of memory, then assembles the relevant records within a token budget. Across three writing benchmarks, NarraWorld achieves the strongest aggregate results. Its memory also transfers to situated role-playing and largely preserves recall on a general-purpose long-term memory benchmark, paving the way for agents that sustain coherent storyworlds across diverse narrative tasks.
\end{abstract}

\etocdepthtag.toc{main}

\section{Introduction}
\label{sec:introduction}
LLM agents are increasingly used to assist long-form fiction writing: drafting
stories \citep{yang2022re3}, expanding outlines into full narratives
\citep{yang2023doc}, and co-authoring with human writers
\citep{mirowski2023dramatron,huot2025agentsroom}. At this scale,
quality depends on narrative consistency, meaning that new events must agree
with facts already established in the story. A story's history can quickly
outgrow a single model context, and even within long contexts, access to
relevant information degrades when it is distant or appears in the middle of
the input \citep{liu2023lostmiddle}. Long-form writing therefore requires an
explicit memory for tracking the evolving story state.

\begin{figure*}[t]
\centering
\includegraphics[width=\textwidth]{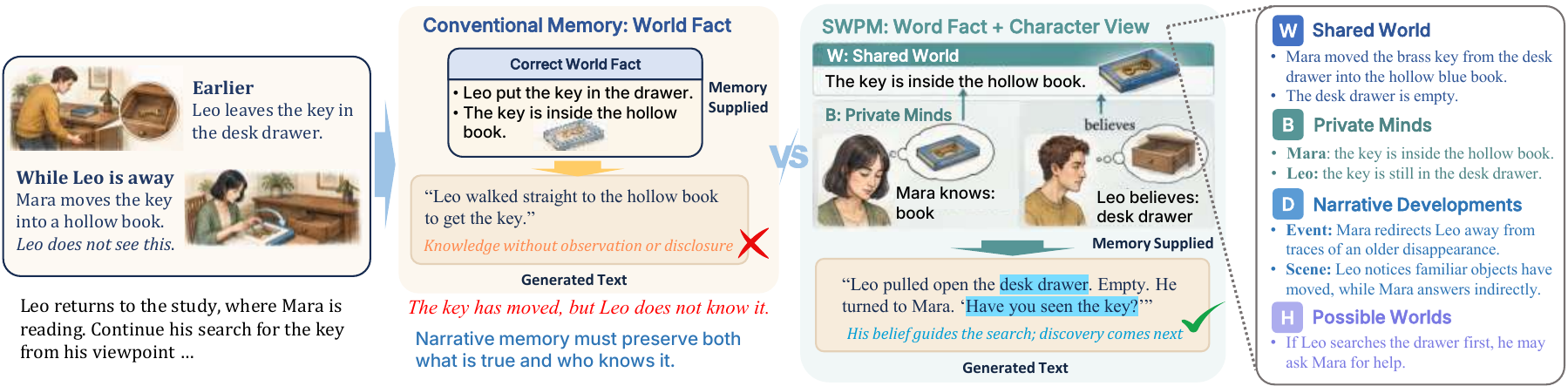}
\caption{Illustrative comparison of conventional memory and NarraWorld. Leo leaves the key in the desk drawer, but Mara moves it into a
hollow book while he is away. Conventional memory supplies the updated location, causing Leo to walk directly to
the book. NarraWorld
preserves both the shared fact that the key has moved and Leo's private belief
that it remains in the drawer, so the continuation is: he
checks the drawer, discovers it empty, and asks Mara about the key.}
\label{fig:illustrative-example}
\end{figure*}

Such memory must represent the latent storyworld rather than only the surface
story text. This poses three challenges for existing agent memory systems.
First, narrative information is heterogeneous: a passage can simultaneously
mix objective world facts, character beliefs, plot developments, and
foreshadowed possibilities. Although these elements belong to the same
storyworld, they may differ or even conflict in ways that are difficult to
detect from the surface text. For example, after one character places a key in
a drawer, another moves it into a hollow book; the first character's belief
that the key remains in the drawer then conflicts with its actual location.
Memory that
collapses these statuses \citep{park2023generative,packer2023memgpt,
zhong2024memorybank,xu2026structmem}
will conflate information that should remain distinct, leading to
contradictions in the story (\textbf{Figure
\ref{fig:illustrative-example}}).

Second, narrative development often accumulates gradually across many passages. For
example, a betrayal may build up over ten chapters through small acts of
concealment and misdirection. Memory must be able to summarize it as a single
betrayal from multiple chapters without losing the specific facts. Integrating
such evidence across multiple narrative granularities challenges existing
methods, whose memories are typically organized as flat summaries, isolated
facts, or event-level records
\citep{piper2021narrative,lyu2025facttrack,wang2025dome}.

Third, a writing query often provides too little information to retrieve all
the memory needed for the task. For example, a simple instruction such as
``rewrite the betrayal scene'' requires not only memories of the betrayal
itself but also the character relationships that the revision may affect. Existing systems
retrieve by similarity \citep{park2023generative,zhong2024memorybank}, so they are bounded by the request text itself
and cannot extend to these implicit dependencies.

To this end, we present NarraWorld, a memory system that treats long-form
writing as world creation.
It maintains four kinds of records: world facts, per-character beliefs, open
developments, and hypothetical branches. These four layers are connected views
derived from a shared evidence-grounded graph, with the belief view indexed by
character.
A hierarchical aggregation mechanism with atomic closure consolidates events
into scenes, plotlines, and plots. Every higher-level node can be traced through
the hierarchy to all of its constituent events and source spans, so a
multi-chapter development can be retrieved as a single unit and expanded back
into its exact underlying facts.
During memory retrieval, a planned reconstruction determines a query's
dependency before retrieving. A planner reads the current narrative situation
then checks this against a preview of memory, derives the dependencies,
retrieves and assembles the related records within a token budget.

Experiments show that across three writing benchmarks, NarraWorld achieves the highest aggregate writing
quality. The same
memory transfers to situated role-playing, which likewise requires
maintaining an underlying storyworld. On a general-purpose long-term memory
task, the same memory largely preserves recall performance.
Our contributions are fourfold:

\noindent(1) We propose a shared evidence-grounded memory substrate that
    separates world facts, per-character beliefs, open developments, and
    hypothetical branches.

\noindent(2) We propose a hierarchical aggregation mechanism with atomic closure
    that organizes events into scenes, plotlines, and plots while preserving
    links to constituent events and source spans.

\noindent(3) We propose a planned reconstruction procedure that derives implicit
    continuation dependencies from the narrative situation before retrieving
    and assembling memory.

\noindent(4) Experiments show that NarraWorld improves long-form consistency and writing quality
while transferring to situated role-playing.

\section{NarraWorld: Structured Memory as World Creation}
\label{sec:method}

\begin{figure*}[t]
\centering
\includegraphics[width=\textwidth]{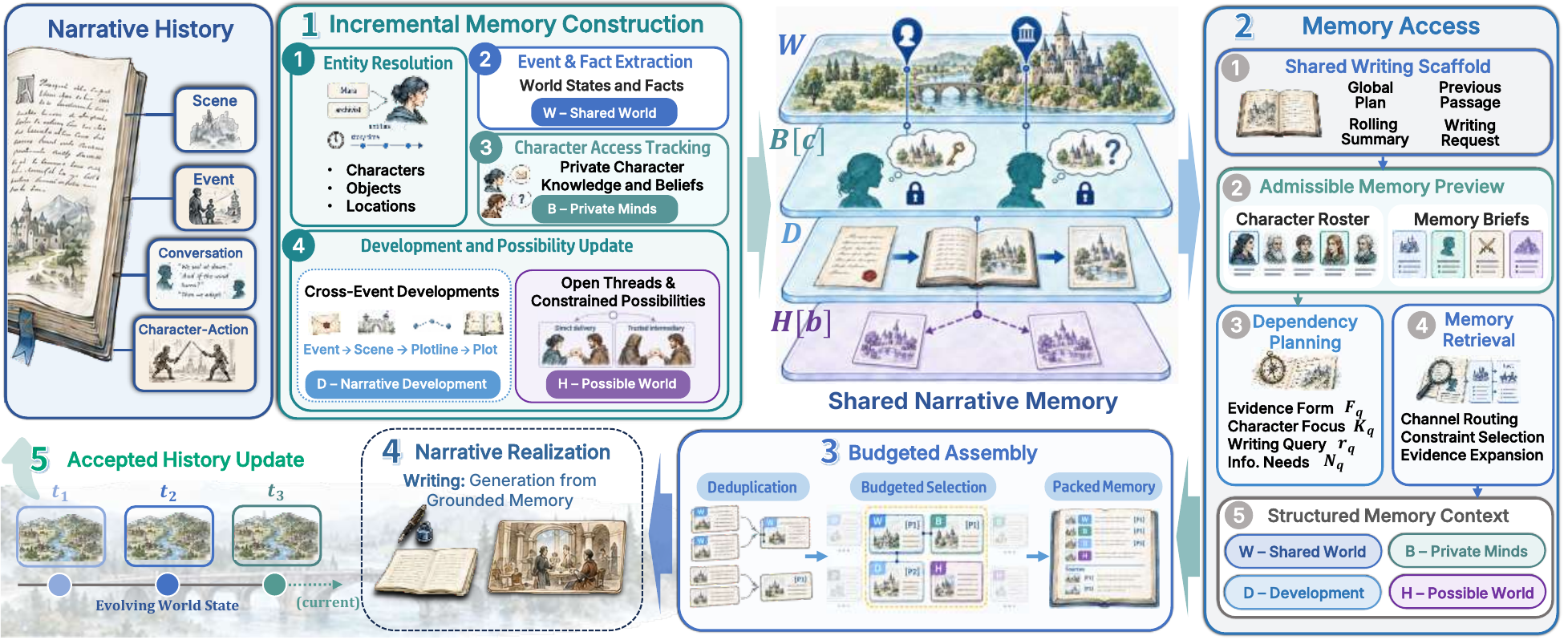}
\caption{Overview of the NarraWorld continual world-creation cycle. Narrative history
is formed into shared-world and character-belief memory, then
reconstructed for the next writing act or situated agent action. The resulting
continuation becomes the next source unit in the evolving narrative.}
\label{fig:overview}
\end{figure*}

As shown in Figure~\ref{fig:overview}, NarraWorld has three components.
First, an evidence-grounded substrate separates world facts, private character
knowledge, narrative developments, and possible worlds. Second, a hierarchy
organizes atomic events into scenes, plotlines, and plots. Atomic closure links
every abstraction to its events and source evidence. Third, planned
reconstruction uses the current narrative situation and a memory preview to
infer implicit dependencies before retrieving and assembling an admissible
context under a budget. The operational construction sequence is detailed in
Appendix~\ref{app:construction-details}.

\subsection{Task Formulation}
\label{sec:problem}

We adopt the plan--write--revise setting of Re3
\citep{yang2022re3}. At writing step $\tau$, the agent receives a scaffold
\begin{equation}
x_\tau=\left(O,\rho_\tau,P_{\tau-1},r_\tau,\kappa_\tau\right),
\label{eq:shared-writing-scaffold}
\end{equation}
where $O$ is the global outline, $\rho_\tau$ a rolling summary,
$P_{\tau-1}$ the preceding passage or chapter, $r_\tau$ the current request,
and $\kappa_\tau$ the output constraints. We represent the complete narrative
request as $q=(x_\tau,\eta_q)$, where $\eta_q$ specifies the teller, recipient,
focalizer, story time, discourse position, and applicable branch. NarraWorld
reconstructs memory for the next writing action from this request.

\subsection{Multi-View Memory over a Shared Narrative Graph}
\label{sec:unified-narrative-graph}
Long-form narrative memory must preserve two properties. World facts,
character beliefs, narrative developments, and hypothetical continuations
require different evidential status, yet they often refer to the same entities,
events, times, and source passages. Narratology similarly treats story,
narrative discourse, and narrating as distinct but interdependent levels
\citep{genette1983narrative}. Inspired by this principle, NarraWorld therefore maintains one evolving narrative structure and forms four connected memory
views over it.
\paragraph{Four views projected from the shared structure.}
At time $\tau$, the shared structure $G_\tau$ connects recurring entities, events, times, and
source evidence across views. The memory $M_\tau$ contains four views: $W_\tau$ represents established world state, including events that
have occurred, currently valid facts, entity states, and relationships.
$B_\tau[c]$ represents the current perspective, such as knowledge and beliefs, of character $c$. The two views refer to the
same events and propositions but assign them different perspective-dependent
status. Since long-form writing also depends on narrative developments (e.g., conflicts, relationships) that develop across
passages and chapters, the development view $D_\tau$ connects the relevant events and state changes
and records the current status of each development, such as open, advanced,
blocked, or resolved. During story writing, an open development may suggest multiple future branches. $H_\tau$ stores bounded hypothetical continuations (possible-world branches) associated
with open developments. All four views are obtained by projecting $G_\tau$.

\begin{equation}
\begin{aligned}
W_\tau &= \Pi^W_\tau(G_\tau), &
B_\tau[c] &= \Pi^{B[c]}_\tau(G_\tau),\\
D_\tau &= \Pi^D_\tau(G_\tau), &
H_\tau &= \Pi^H_\tau(G_\tau).
\end{aligned}
\label{eq:semantic-view-projections}
\end{equation}

Each projection starts from a different set of nodes and expands as a different subgraph in $G_\tau$. The world projection $\Pi^W_\tau$ contains
source-supported events, facts, states, and relationships that are valid at
time $\tau$. The character projection $\Pi^{B[c]}_\tau$ starts from character
$c$ and follows action and belief links (e.g., observation and communication) to propositions held by the character, which may be either world facts or false propositions that contradict them. The development projection
$\Pi^D_\tau$ starts from nodes associated with narrative development, such as goals and conflicts, and organizes events through progression, causation, and state change (see Section \ref{sec:narrative-hierarchy}). The possibility projection $\Pi^H_\tau$ starts from unresolved
developments and collects their constrained candidate continuations and marks them with hypothetical status.

\paragraph{Incremental memory construction.}
During the writing process, given a new unit $U_\tau$ of the source text, NarraWorld updates the shared narrative structure:
\begin{equation}
G_\tau=\operatorname{Update}(G_{\tau-1},U_\tau).
\label{eq:shared-structure-update}
\end{equation}
The update contains four stages. First, new character, location, and object entities in $U_\tau$ are linked to existing entities; unmatched entities create new records. Second, using these resolved entities, NarraWorld adds new events together with the facts and state changes supported by those events. When an event changes an existing state, NarraWorld preserves the validity period of the previous state and establishes a new current state instead of overwriting the earlier record. Third, NarraWorld determines each character's access to the established events and facts. When a character observes an event, receives information, or draws an inference, NarraWorld connects that character to the corresponding event or proposition and records the acquisition mode. Fourth, NarraWorld updates narrative developments from the new events and state changes. For developments that remain unresolved, it generates conditional candidates constrained by the current world state and character knowledge and adds them to $H_\tau$.

\paragraph{An example for multi-view update.}
For the source text ``Mara moved the key from the drawer into the hollow book
while Leo was away,'' NarraWorld records the movement event and updates the key's
location. Because the event and resulting location are source-supported and
current, $\Pi^W_\tau$ includes them in $W_\tau$. Mara participates in the event,
so her access link allows $\Pi^{B[\text{Mara}]}_\tau$ to include the new
location. Leo has no observation or communication link to the move, so
$\Pi^{B[\text{Leo}]}_\tau$ retains his earlier belief that the key is in the
drawer. The movement event advances the development concerning Leo's search;
$\Pi^D_\tau$ therefore includes the event under that development. If the
development remains unresolved, $\Pi^H_\tau$ retrieves candidates such as Mara
informing Leo or Leo discovering the key himself.

\subsection{Organizing Memory across Narrative Scales}
\label{sec:narrative-hierarchy}

NarraWorld organizes narrative memory from events to scenes, plotlines, and plots.
This hierarchy connects local evidence with narrative developments that
accumulate across the story:
\begin{equation}
\mathcal{E}_\tau
\longrightarrow
\mathcal{S}_\tau
\longrightarrow
\mathcal{L}_\tau
\longrightarrow
\mathcal{P}_\tau,
\label{eq:narrative-hierarchy}
\end{equation}
corresponding to events, scenes, plotlines, and plots.

$\mathcal{E}_\tau$ consists directly of event records in canonical narrative
memory. Each event forms an atomic node and retains its participants, temporal
scope, source unit, and supporting evidence.
$\mathcal{S}_\tau$ consists of derived scene bundles. Each bundle gathers the
events, state changes, relationship evolution, and plan progress within one
local narrative unit and summarizes them as a scene.
$\mathcal{L}_\tau$ identifies persistent goals, conflicts, relationships, and
plans across scenes. Candidate continuity comes from shared characters,
continuous content, state transitions, developmental relations, and thread
evolution. Given the selected scenes, an LLM produces a plotline summary and
the connections among its scenes, with explicit references to the supporting
scene records.
$\mathcal{P}_\tau$ combines plotlines that share participants and exhibit common
continuity or developmental signals. The resulting plot forms a connected
narrative arc and aggregates its member plotline summaries in temporal order.

Each higher-level node retains an atomic closure of its member events, $W/B/D$
records, and supporting source spans. The hierarchy can therefore represent a
multi-scene development as one plotline or plot while preserving every lower-level
record from which that representation was derived. Retrieval can select the
higher-level unit and then unfold it into the relevant actions, beliefs, and
source evidence. These derived views evolve alongside atomic memory.
Hypothetical $H$ records remain outside the hierarchy until subsequent source
text realizes them as $W/B/D$ evidence.

\subsection{Reconstructing Memory for Situated Writing}
\label{sec:online-reconstruction}

Writing requests are underspecified retrieval queries. A request may name an
event but omit affected relationships, character knowledge boundaries, and
unresolved consequences. NarraWorld reconstructs memory in four stages: admissible
view projection, dependency planning, multi-scale retrieval, and budgeted
assembly.

\paragraph{Situation-conditioned memory access.}
Relevant memory and usable memory differ by story time, disclosure state,
focalization, and branch. For example, in a mystery, the detective may not yet
know the truth, while the culprit's identity should not be disclosed
prematurely. The narrative situation $\eta_q$ defines these access
conditions and situates memory in an occasion of telling
\citep{herman2009basic,piper2021narrative}. It determines the admissible view
\begin{equation}
V_q=\Pi_{\eta_q}(\mathcal{M}_{\tau}),
\label{eq:admissible-view}
\end{equation}
where $\Pi$ identifies records available under the request's temporal,
disclosure, and branch scope. It preserves record authority and belief
holders. The author-facing view can include beliefs from several characters.
Holder labels preserve their separation. Focalization determines which
character's available information may be expressed in the narration.

\paragraph{Preview-guided dependency planning.}
Similarity search follows explicit query content. Implicit dependencies require
an intermediate planning step.
Appendix~\ref{app:retrieval-ablation} compares this preview-conditioned design
with deterministic and model-based query-centric retrieval on LoCoMo, and with
scaffold-conditioned query planning on ConStory-Bench.
The planner receives the request and $\sigma(V_q)$, a compact preview containing
the full character roster, brief descriptions, developments, and $W/B/D/H$
records. The current passage and authorial brief provide immediate context.
Outlines provide intended directions. Planning produces
\begin{equation}
(F_q,K_q,r_q,\mathcal{N}_q)
=\Psi(x_\tau,\eta_q;\sigma(V_q)),
\label{eq:information-needs}
\end{equation}
where $F_q$ specifies the required evidence form and composition, $K_q$
identifies focal characters, $r_q$ states the content-focused retrieval intent,
and $\mathcal{N}_q$ lists missing information needs and their dependencies.
The system resolves $K_q$ against the admissible character view and translates
these semantic decisions into retrieval constraints. This character-centered
planning can recover relevant entities even when they are absent from the
surface request.

\paragraph{Scale-adaptive retrieval.}
Narrative dependencies map to different memory units: facts to atomic records,
local interactions to events and scenes, and ongoing developments to scenes and
plotlines.
Retrieval uses these planning outputs to expand the admissible candidate set and
search for records that satisfy the remaining needs:
\begin{equation}
E_q
=
\operatorname{Expand}(K_q,F_q;V_q)
\cup\mathcal{R}(x_\tau,r_q,\mathcal{N}_q,K_q,F_q;V_q).
\label{eq:narrative-retrieval}
\end{equation}
Expansion follows the focal characters and requested evidence composition to
construct an admissible candidate set; search then satisfies the remaining
needs. It combines lexical and semantic matching with temporal, epistemic,
and developmental access.
Higher-level matches unfold into their supporting records.
The retrieval uses holder and time to distinguish world propositions,
character stances, and acquisition events.

\paragraph{Budgeted assembly.}
Given a memory budget $b$, NarraWorld renders the retrieved evidence
as a readable context
\begin{equation}
C_q^{\mathrm{mem}}=\operatorname{Reconstruct}(q,V_q;b),
\qquad \ell(C_q^{\mathrm{mem}})\leq b,
\label{eq:budgeted-reconstruction}
\end{equation}
where $\ell$ measures rendered length. Evidence is grouped into $W/B/D/H$
sections, with $B$ organized by character. Hypothetical candidates retain their
premises and conditions without presenting possibilities as accepted evidence.
The writer agent receives
$X_q=x_\tau\oplus C_q^{\mathrm{mem}}$ and uses it to produce the next passage.
Before history exists, the scaffold supplies the authorial
brief and constraints; subsequent memory updates use only accepted source text.

\section{Experiments}
\label{sec:experiments}

\subsection{Experimental Setup}
\label{sec:experimental-setup}

\paragraph{Tasks and evaluation data.}
Our primary evaluation covers three writing benchmarks: ConStory-Bench has 200
stories equally divided among generation, continuation, expansion, and
completion \citep{li2026constory}; TMAS has 168 source-grounded
continuation checkpoints \citep{huot2025agentsroom}; and LSGC has five official
starters, each extended to 14 chapters and at least 40,000 words
\citep{migal2024lsgc}. We retain the
original writing tasks while adapting generation into incremental segments that
better reflect writers' workflows and expose memory use across successive
writing steps. For LSGC, every condition is evaluated in three complete runs;
the reported scores average the resulting 15 story instances, each first
aggregated over its 14 chapter judgments and final-piece judgment. We also use RMTBench's 2,880 multi-turn role-playing trajectories
to assess whether NarraWorld transfers beyond writing \citep{xiang2025rmtbench}, and
LoCoMo's 1,540 main recall questions to test whether its writing-oriented design
compromises general-purpose memory performance \citep{maharana2024locomo}.
Appendix~\ref{app:writing-adaptations} details the writing protocols.

\paragraph{Baselines and model configuration.}
We compare NarraWorld with Full Context, Scaffold Only, Mem0
\citep{chhikara2025mem0}, Graphiti \citep{rasmussen2025zep}, Hindsight
\citep{latimer2026hindsight}, AriGraph \citep{anokhin2025arigraph}, and StructMem
\citep{xu2026structmem}, adding MemoryOS \citep{kang2025memoryos} for LoCoMo
and Emotional RAG for RMTBench. Full Context gives the TMAS writer the visible
source prefix directly; Scaffold Only omits retrieved memory from the shared
writing scaffold. All benchmarks use GPT-5-mini for memory-system calls, GPT-5
for generation, and DeepSeek-V4-Pro for judging. Systems retain their native
memory procedures, with embedding and reranking services fixed by benchmark.
Rendered-memory budgets are 12k tokens for ConStory-Bench and LSGC, 4k for
RMTBench, and 8k for LoCoMo (2k--16k in the sensitivity analysis).
Throughout the paper, ``Mem. tokens'' denotes the average rendered memory length
under the corresponding budget.
Conditions share task inputs, writing models, and evaluation protocols, and
exclude future text and evaluator-only references.

\paragraph{Metrics and evaluation protocol.}
ConStory-Bench reports Consistency Error Density (CED) across five
categories and 19 subtypes, plus Words per Penalized Error (WPE),
the story-level word count divided by one plus the subtype count
\citep{li2026constory}. TMAS reports four primary dimensions---Plot,
Creativity, Development, and Language Use---following its source-grounded
continuation setting \citep{huot2025agentsroom}. We also report their
arithmetic mean, TMAS Quality, as a supplementary aggregate;
auxiliary diagnostics appear in Appendix~\ref{app:tmas-adaptation}.
Because the original LSGC evaluation relies heavily on human annotation, we use
the 16 Longform Writing Bench criteria for scalable quality evaluation
\citep{paech2025longform},
reporting direction-corrected 0--100 scores for Character,
Plot, World/Coherence, Style/Creativity, and Prompt Adherence; Overall equally
weights all criteria. RMTBench reports Emotional Expression (EE), Emotional
Comprehension (EC), Plot Advancement (PA), Character Understanding (CU),
Character Maintenance (CM), and User Preference Awareness (UPA)
\citep{xiang2025rmtbench}. LoCoMo reports Overall Main and single-hop,
multi-hop, temporal, and open-domain scores \citep{maharana2024locomo}.
Appendix~\ref{app:evaluation-protocols} provides their definitions.

\subsection{Overall Results}
\label{sec:results}

\paragraph{Quantitative results.}
Across all three writing benchmarks, NarraWorld provides the strongest overall
results (Tables~\ref{tab:constory-results}--\ref{tab:tmas-results}). It leads
all six reported ConStory-Bench measures, including CED of $0.185$
versus $0.200$ for Full Context and $0.385$ for Graphiti, as well as the
highest WPE. On LSGC, averaged over three runs, it leads Character ($91.12$), Plot ($90.50$),
Style/Creativity ($88.97$), and Overall ($90.24$), ahead of StructMem
($86.92$) and Full Context ($85.03$); the full rubric shows a remaining Full
Context advantage on World/Coherence (Appendix~\ref{tab:lsgc-criteria}).
Run-level ranges and sample standard deviations are reported in
Appendix~\ref{tab:lsgc-run-variability}.
NarraWorld also leads 13 of the 16 LSGC criteria and the TMAS Creativity
($84.90$), Development ($85.60$), and aggregate Quality measures, while Full
Context leads TMAS Plot ($80.03$) and Language ($83.42$). Graphiti uses more
rendered memory than NarraWorld on both ConStory-Bench ($5,971$ vs. $3,231$)
and LSGC ($10,542$ vs. $6,810$), yet remains weaker on consistency and
long-form quality, indicating that more connected history is insufficient
without perspective and development modeling. NarraWorld's gains over the
strongest LSGC memory baseline, StructMem, further support narrative-specific
views and hierarchical organization. Full Context remains a strong reference,
but exposing complete history at every step is difficult for very long
iterative writing and is not a practical deployment strategy.

\begin{table*}[t]
\caption{ConStory-Bench consistency results. CED and task-specific errors are
lower-is-better; WPE is an auxiliary higher-is-better word-per-error measure.
}
\label{tab:constory-results}
\centering
\small
\setlength{\tabcolsep}{2.8pt}
\begin{tabular*}{\textwidth}{@{\extracolsep{\fill}}lrrrrrrr@{}}
\toprule
Method & CED $\downarrow$ & WPE $\uparrow$ & Generation & Continuation & Expansion & Completion & Mem. tokens \\
\midrule
NarraWorld & \textbf{0.185} & \textbf{10,520.4} & \textbf{0.120} & \textbf{0.235} & \textbf{0.098} & \textbf{0.287} & 3,231 \\
Graphiti & 0.385 & 9,480.5 & 0.290 & 0.415 & 0.310 & 0.525 & 5,971 \\
StructMem & 0.450 & 9,215.3 & 0.355 & 0.495 & 0.380 & 0.570 & 3,891 \\
Mem0 & 0.955 & 7,845.2 & 0.812 & 0.985 & 0.920 & 1.103 & 2,179 \\
AriGraph & 1.125 & 7,250.6 & 0.945 & 1.180 & 1.055 & 1.320 & 3,050 \\
Hindsight & 1.215 & 6,985.4 & 1.050 & 1.265 & 1.140 & 1.405 & 3,280 \\
Scaffold Only & 1.350 & -- & 1.220 & 1.410 & 1.280 & 1.490 & -- \\
\midrule
Full Context & 0.200 & 9,847.8 & 0.136 & 0.255 & 0.113 & 0.298 & -- \\
\bottomrule
\end{tabular*}
\end{table*}

\begin{table*}[t]
\begin{minipage}[t]{0.49\textwidth}
\centering
\vspace{0pt}
\small
\captionof{table}{LSGC long-form writing results ($\uparrow$). Direction-corrected
macro means over five starters and three runs; memory in tokens.}
\label{tab:lsgc-results}
\setlength{\tabcolsep}{1.5pt}
\footnotesize
\begin{tabular*}{\linewidth}{@{\extracolsep{\fill}}lrrrrr@{}}
\toprule
Method & Character & Plot & \shortstack{Style/\\Creat.} & Overall & \shortstack{Mem.\\tokens} \\
\midrule
NarraWorld & \textbf{91.12} & \textbf{90.50} & \textbf{88.97} & \textbf{90.24} & 6,810 \\
Mem0 & 84.34 & 81.71 & 73.69 & 80.18 & 2,876 \\
Graphiti & 82.44 & 87.11 & 67.34 & 79.10 & 10,542 \\
Hindsight & 75.32 & 79.46 & 67.73 & 74.91 & 3,412 \\
AriGraph & 72.56 & 76.87 & 67.56 & 73.03 & 3,850 \\
StructMem & 88.87 & 87.74 & 83.60 & 86.92 & 5,428 \\
Scaffold Only & 66.50 & 69.22 & 70.86 & 69.61 & -- \\
Full Context & 87.17 & 87.27 & 78.79 & 85.03 & -- \\
\bottomrule
\end{tabular*}
\end{minipage}\hfill
\begin{minipage}[t]{0.49\textwidth}
\centering
\vspace{0pt}
\small
\captionof{table}{TMAS continuation results ($\uparrow$). Primary dimensions
are macro means; more diagnostic metrics are reported in Appendix
Table~\ref{tab:tmas-auxiliary}.}
\label{tab:tmas-results}
\setlength{\tabcolsep}{1.5pt}
\footnotesize
\begin{tabular*}{\linewidth}{@{\extracolsep{\fill}}lrrrrr@{}}
\toprule
Method & Plot & \shortstack{Creat.} & \shortstack{Develop-\\ment} & \shortstack{Lang.} & \shortstack{Mem.\\tokens} \\
\midrule
NarraWorld & 78.92 & \textbf{84.90} & \textbf{85.60} & 82.62 & 1,593 \\
Mem0 & 77.60 & 84.27 & 82.19 & 83.08 & 941 \\
Graphiti & 79.13 & 83.65 & 82.40 & 82.92 & 1,873 \\
Hindsight & 75.58 & 79.65 & 76.83 & 79.42 & 1,481 \\
AriGraph & 76.00 & 77.56 & 75.79 & 77.75 & 981 \\
StructMem & 77.71 & 84.06 & 82.25 & 82.17 & 1,927 \\
Scaffold Only & 77.10 & 81.50 & 80.90 & 82.20 & -- \\
Full Context & \textbf{80.03} & 83.33 & 83.73 & \textbf{83.42} & -- \\
\bottomrule
\end{tabular*}
\end{minipage}
\end{table*}

\paragraph{Human evaluation.}
Human readers support the main writing conclusion: on ConStory-Bench, NarraWorld
and Full Context have nearly identical human CED (0.095 and 0.096), both below
Graphiti (0.289). For LSGC, readers evaluated complete stories across five
quality dimensions.
NarraWorld obtained the highest LSGC Human Overall (84.50 versus 80.50 for StructMem 
and 80.06 for Full Context). Judge--human agreement was moderate overall
(Spearman $\rho=0.55$), with reader variation in subjective literary judgments;
evaluation protocols, alignment details, and human rubrics appear in
Appendix~\ref{app:human-evaluation}.

\subsection{Ablation Studies}
\label{sec:ablations}

\begin{figure*}[!htbp]
\centering
\begin{minipage}[t]{0.6\textwidth}
\centering
\vspace{0pt}
\includegraphics[width=\linewidth]{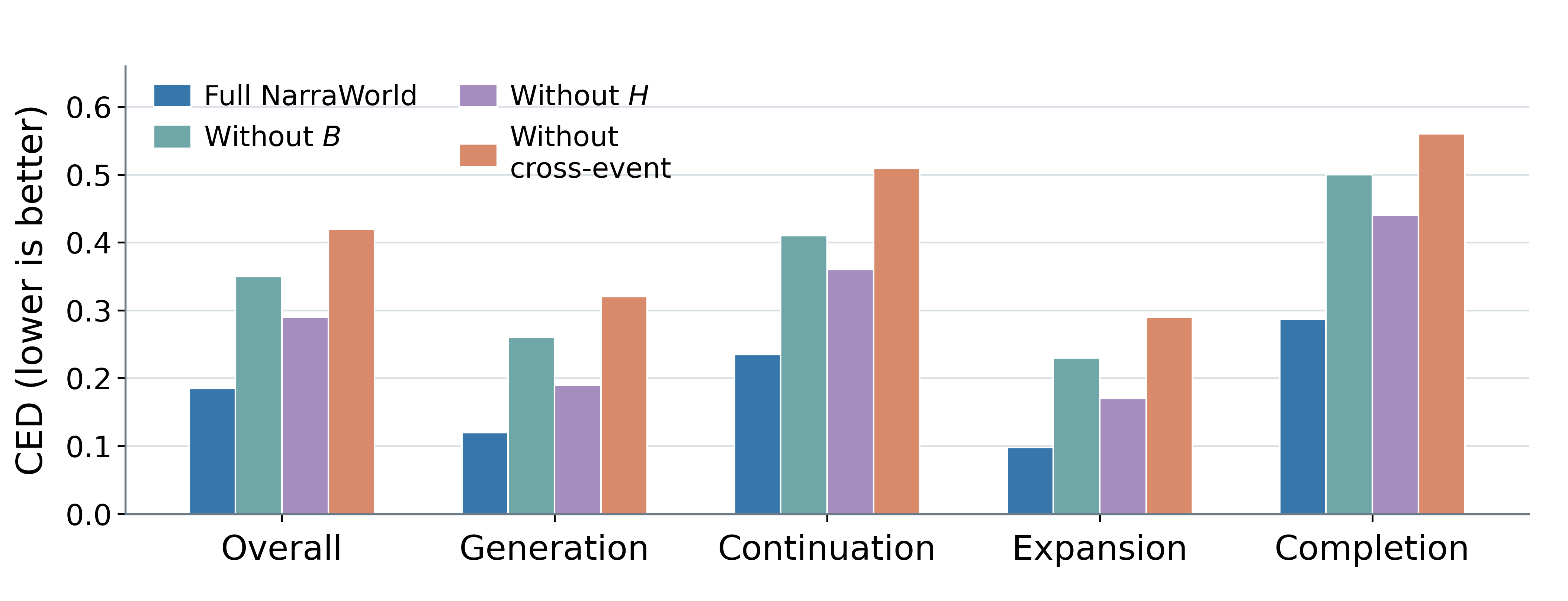}
\captionof{figure}{ConStory-Bench ablations across task types. CED is
lower-is-better. The four settings are Full NarraWorld, without character beliefs
($B$), without possible worlds ($H$), and without cross-event organization.}
\label{fig:constory-ablation}
\end{minipage}\hfill
\begin{minipage}[t]{0.4\textwidth}
\centering
\vspace{0pt}
\scriptsize
\captionof{table}{LSGC ablation Overall scores ($\uparrow$), averaged over
five starters and three runs.}
\label{tab:lsgc-core-ablations}
\setlength{\tabcolsep}{2.5pt}
\begin{tabular}{@{}lr@{}}
\toprule
Setting & Overall $\uparrow$ \\
\midrule
Full NarraWorld & \textbf{90.24} \\
Without character beliefs ($B$) & 86.04 \\
Without possible worlds ($H$) & 88.74 \\
Without cross-event organization & 81.05 \\
\bottomrule
\end{tabular}
\end{minipage}
\end{figure*}

The component ablations show distinct, nonuniform contributions (Figure~\ref{fig:constory-ablation};
Table~\ref{tab:lsgc-core-ablations}). Removing character beliefs weakens
perspective-sensitive continuity, while removing possible worlds lowers
creativity and development; removing cross-event organization causes the
broadest degradation across consistency and long-form quality.
The largest decline occurs when cross-event organization is removed, suggesting
that retaining individual facts is insufficient without connecting them into
sustained narrative developments. The retrieval-planning ablation holds offline
memory and budget fixed:
preview-guided planning improves every ConStory-Bench task, with the largest
gains on continuation and completion. Complete breakdowns appear in
Appendix~\ref{app:memory-ablations}, and retrieval comparisons in
Appendix~\ref{app:retrieval-ablation}.

\subsection{Case Studies}
\label{sec:case-studies}

The case studies show how NarraWorld supports writing-specific retrieval. In
the selected ConStory-Bench continuation, it jointly retrieves an unfinished
fox drawing, the narrator's knowledge of a condition attached to their
mother's box, and the father's promise to give them space. The continuation
completes the drawing while the father remains outside and refrains from asking
to see it, combining a factual callback with private access and relational
commitment. The full comparison appears in Appendix~\ref{app:qualitative-constory};
further cases appear in Appendix~\ref{app:qualitative-additional}.

\subsection{Extension to Situated Role-Playing}
\label{sec:role-playing}
Because NarraWorld reconstructs the storyworld, it also supports role-playing
tasks that depend on character beliefs. On RMTBench it leads all four 1--5
dimensions, including plot advancement and character understanding
(Table~\ref{tab:rmt-results}); Full Context is slightly higher on the
percentage-based CM and UPA measures. This transfer shows that separating
shared events from character-specific access helps an agent preserve its role
while extending the situation.

\begin{table*}[t]
\caption{RMTBench multi-turn role-playing results ($\uparrow$). EE/EC/PA/CU use
1--5 scores; CM/UPA are percentage success rates. Scores are averaged over
applicable turns.}
\label{tab:rmt-results}
\centering
\small
\setlength{\tabcolsep}{4pt}
\begin{tabular}{lrrrrrrr}
\toprule
Method & EE $\uparrow$ & EC $\uparrow$ & PA $\uparrow$ & CU $\uparrow$ &
CM $\uparrow$ & UPA $\uparrow$ & Mem. tokens \\
\midrule
NarraWorld & \textbf{4.22} & \textbf{4.06} & \textbf{4.30} & \textbf{4.57} & 87.15\% & 68.85\% & 704 \\
Mem0 & 3.98 & 3.56 & 3.74 & 4.25 & 85.00\% & 67.31\% & 344 \\
Graphiti & 4.02 & 3.61 & 3.82 & 4.41 & 86.15\% & 53.85\% & 1,215 \\
Hindsight & 4.01 & 3.52 & 3.78 & 4.21 & 84.60\% & 65.40\% & 654 \\
AriGraph & 3.95 & 3.60 & 3.71 & 4.28 & 85.35\% & 68.10\% & 429 \\
Emotional RAG & 3.99 & 3.55 & 3.73 & 4.22 & 82.10\% & 67.30\% & 600 \\
StructMem & 3.96 & 3.49 & 3.67 & 4.38 & 85.71\% & 50.00\% & 2,357 \\
Full Context & 3.92 & 3.52 & 3.66 & 4.30 & \textbf{87.86\%} & \textbf{69.23\%} & -- \\
\bottomrule
\end{tabular}
\end{table*}

\subsection{General-Purpose Memory}
\label{sec:general-memory}

\begin{figure*}[t]
\centering
\begin{minipage}[t]{0.42\textwidth}
\centering
\vspace{0pt}
\includegraphics[width=\linewidth]{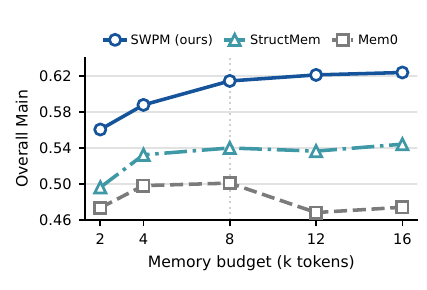}
\captionof{figure}{LoCoMo budget sensitivity. The vertical guide marks
the 8k budget used in the main comparison.}
\label{fig:locomo-budget}
\end{minipage}\hfill
\begin{minipage}[t]{0.55\textwidth}
\centering
\vspace{0pt}
\captionof{table}{LoCoMo results at the shared 8k memory budget. Overall Main
aggregates the four main question types ($\uparrow$).}
\label{tab:locomo-results}
\scriptsize
\setlength{\tabcolsep}{2pt}
\begin{tabular*}{\linewidth}{@{\extracolsep{\fill}}lrrrrr@{}}
\toprule
Method & \shortstack{Overall\\Main} & \shortstack{Multi-\\hop} & \shortstack{Single-\\hop} & Temporal & \shortstack{Open-\\domain} \\
\midrule
NarraWorld & \textbf{0.614} & \textbf{0.458} & 0.679 & \textbf{0.649} & \textbf{0.392} \\
Graphiti & 0.582 & 0.444 & \textbf{0.698} & 0.469 & 0.351 \\
StructMem & 0.540 & 0.435 & 0.575 & 0.595 & 0.362 \\
Mem0 & 0.501 & 0.410 & 0.541 & 0.547 & 0.270 \\
AriGraph & 0.495 & 0.378 & 0.623 & 0.314 & 0.322 \\
Hindsight & 0.486 & \textbf{0.458} & 0.556 & 0.394 & 0.264 \\
MemoryOS & 0.453 & 0.371 & 0.537 & 0.354 & 0.292 \\
\bottomrule
\end{tabular*}
\end{minipage}
\end{figure*}

To test transfer beyond writing, we evaluate NarraWorld on LoCoMo. Under the
shared main budget, it leads or ties four of five reported measures, with
Graphiti ahead only on single-hop questions (Table~\ref{tab:locomo-results});
NarraWorld also ranks first throughout the budget sweep
(Figure~\ref{fig:locomo-budget}). Complete category-level results appear in
Appendix~\ref{app:locomo-metrics}. For this QA task, the default memory view
excludes hypothetical-world records ($H$): exposing that view lowers Overall
Main from 0.614 to 0.570 at the same 8k budget (Appendix~\ref{tab:locomo-full-ablation}).
This contrast suggests that factual recall and creative continuation require
different views of the same memory: hypothetical alternatives can support
writing, while QA benefits from established evidence.

\section{Related Work}
\label{sec:related-work}

\paragraph{Long-form Narrative Generation and Storyworld Modeling}

Long-form writing systems use hierarchical outlines, recursive drafting and
revision, multi-agent planning, or explicit story-state models. These families
include retrieved-summary revision and detailed outline control
\citep{yang2022re3,yang2023doc}, structured co-writing roles
\citep{mirowski2023dramatron,huot2025agentsroom}, and temporal, validity-aware,
or character-grounded world models
\citep{wang2025dome,lyu2025facttrack,aluru2026magnet}. They establish the
need for state that persists beyond the model context. Most evaluations still
produce one complete story or continuation from a fixed prompt, prefix, or
outline. NarraWorld instead treats writing as a sequence of accepted narrative acts:
the agent generates one segment, updates the story state, and reconstructs
memory for the next request. This makes memory responsible for consequences,
perspective, and unresolved developments across steps.

\paragraph{Computational Narratology and Situated Character Minds}

Computational narratology distinguishes story events from their discourse-level
selection and ordering, and narratorial voice from the consciousness through
which events are perceived \citep{genette1983narrative,bal2009narratology,
piper2021narrative,herman2009basic}. Work on focalization, imperfect
knowledge, false belief, and narrative time formalizes the associated
perspective constraints \citep{bae2011focalization,teutenberg2015knowledge,
sanghrajka2022headspace,kearns2020narrativetime,han2022flashback}. These
distinctions motivate NarraWorld's four views: shared world state (W), private
character beliefs (B), cross-event development (D), and retained possibilities
(H), the last grounded in possible-world accounts of narrative
\citep{ryan2013possibleworlds}.

\paragraph{Persistent Memory for Situated Language-Model Agents}

Persistent-agent memories support storage, retrieval, reflection, and revision.
Memory streams, working/external memory, and personalized retention represent
three common designs \citep{park2023generative,packer2023memgpt,
zhong2024memorybank}; graph and structured systems add associative traversal,
event binding, and cross-event consolidation
\citep{ji2026reconstructed,xu2026structmem}. Representative query-centric
baselines span flat retained-memory stores (Mem0), temporal or entity-event
graphs (Graphiti and AriGraph), and typed or multi-store organizations
(Hindsight and MemoryOS)
\citep{chhikara2025mem0,rasmussen2025zep,latimer2026hindsight,
anokhin2025arigraph,kang2025memoryos}. These systems improve recall through
query-matched records, but do not jointly model canonical facts, private
knowledge, cross-event development, and possibilities. StructMem binds events
but lacks epistemic and hypothetical views; graph systems do not resolve
canonicity or private belief.
NarraWorld addresses this gap with a shared substrate and task-conditioned
reconstruction for writing and situated action.

\section{Conclusion}
\label{sec:conclusion}

We introduced NarraWorld, a narrative memory system framing long-form writing as 
world creation. It maintains shared world state, private character knowledge, 
cross-event development, and retained possible worlds within a source-grounded 
substrate. It then reconstructs request-specific views constrained by perspective, 
time, and branch. Across ConStory-Bench, Tell Me A Story, and LSGC, NarraWorld 
achieved the strongest aggregate writing results, with ablations showing
complementary but nonuniform contributions from character beliefs, possible
worlds, and cross-event organization. The same representation also supported
situated character action and general-purpose memory use, while the LoCoMo
ablation shows that the admissible memory view should be conditioned on the
task. These results position narrative memory as the reconstruction of a
dynamic, perspective-dependent world rather than relevance-based retrieval.
The remaining challenges are the construction and planning cost of the richer
representation, and deciding when speculative branches should be exposed;
addressing these trade-offs in broader interactive settings is left for future
work.

\clearpage
\subsection*{AI use statement}
Generative AI models perform memory construction, dependency planning,
narrative generation, and evaluation under the reported protocols. AI
assistants also supported conceptual and methodological refinement,
implementation, experiment execution and analysis, literature review, figure
preparation, and manuscript
drafting and editing. The authors reviewed the resulting text, code, and
analyses, checked reported results and citations, and take responsibility for
the final manuscript and research artifacts.

\subsection*{Reproducibility statement}
Section~\ref{sec:method} specifies the memory representation, update process,
hierarchical organization, and online reconstruction procedure, while
Section~\ref{sec:experiments} reports the benchmark scope, model roles, memory
budgets, and comparison controls. Appendix~\ref{app:writing-adaptations}
documents benchmark-specific inputs and adaptations;
Appendix~\ref{app:evaluation-protocols} defines the metrics, aggregation rules,
and evaluation prompts; Appendices~\ref{app:retrieval-ablation} and
\ref{app:memory-ablations} give the ablation protocols; and
Appendices~\ref{app:human-evaluation} and \ref{app:selected-prompts} provide the
human-evaluation procedure and selected system prompts. We retain model
configurations, prompts, per-example outputs, and scoring procedures to support
reproduction, including for models accessed through proprietary APIs. Code and
supplementary materials will be released in anonymized form for review and
made publicly available upon publication.

\subsection*{Ethics statement}
This work uses publicly available benchmark materials and generated fictional
narratives. The human-evaluation analysis reports only anonymized aggregate
judgments and does not expose personally identifiable or sensitive information.
Evaluation materials and records are handled in accordance with the applicable
requirements for the evaluation setting.

\bibliography{references}
\bibliographystyle{iclr2027_conference}

\clearpage
\appendix
\etocdepthtag.toc{appendix}
\etocsettagdepth{main}{none}
\etocsettagdepth{appendix}{subsection}
\renewcommand{\contentsname}{Appendix Contents}
\begingroup
\small
\tableofcontents
\endgroup
\clearpage

\section{NarraWorld Memory Construction}
\label{app:construction-details}

This section expands the construction procedure in Section~\ref{sec:method}.
Each accepted source unit is parsed once, integrated with the existing narrative
structure, and made available through four memory views.
Section~\ref{app:selected-prompts} provides abridged modeling and planning
instructions.

\subsection{Source units and canonical records}
\label{app:construction-records}

The input to an update is a trusted source unit $U_\tau$, consisting of the
newly accepted narrative text together with its source order, optional time
anchor, participants, discourse position, and branch or authority metadata.
The system preserves exact source spans and segments the unit at verified
boundaries. The language-model stages operate on this unit and its local
evidence, while persistent identifiers and source references are assigned by
the system rather than predicted by the model.

The first stage grounds entity mentions against the committed entity index.
It links aliases to existing entities when the evidence supports the match and
creates a new entity otherwise. The next stage extracts events and
propositions: an event records its participants, preconditions, effects, and
causes, while a proposition records its subject, predicate, object, polarity,
and truth status. Each proposition remains independently retrievable. A third
stage models perspective and access, distinguishing a proposition that holds
in the shared world from information that a particular character has observed,
received, or inferred. The fourth stage emits state, relation, and development
deltas. State deltas cover properties such as location, possession, goal,
emotion, and obligation; development deltas mark a thread as opened, advanced,
blocked, or resolved.

After extraction, the system verifies local references, restores exact evidence
spans, and reconciles new entities and propositions with historical records.
Only records with resolved references and retained provenance are committed to
the append-only store. A state change closes the previous validity interval and
creates a new record; it does not overwrite the historical observation. Thus,
the store retains both what was true earlier and what is currently supported.

\subsection{Incremental update and view projection}
\label{app:construction-update}

Let $G_{\tau-1}$ be the committed narrative structure before a new unit. The
update can be summarized as
\begin{equation}
G_{\tau-1}+U_\tau
\;\longrightarrow\;
\text{extract}
\;\longrightarrow\;
\text{reconcile}
\;\longrightarrow\;
\text{commit}
\;\longrightarrow\;
\{W_\tau,B_\tau,D_\tau,H_\tau\}.
\label{eq:appendix-incremental-build}
\end{equation}
The four views are projections of the same committed records, not four
independent memories. $W_\tau$ exposes source-supported world facts, events,
states, and relations. $B_\tau[c]$ follows the access links of character $c$
and may therefore contain a false belief or an incomplete account of the world.
$D_\tau$ links events and state changes to the developments they advance or
block. For an unresolved development, $H_\tau$ stores bounded conditional
continuations whose preconditions are checked against the current world and
character views. A record can participate in several projections while
retaining one provenance chain.

Before each writing or question-answering step, the system projects the
records supported by the available narrative prefix, together with their
derived nodes. It retrieves from this admissible view under the current
temporal, perspective, and branch constraints.

\subsection{Hierarchical organization and evidence closure}
\label{app:construction-hierarchy}

Hierarchical nodes are derived after canonical records are committed. Events
are direct canonical records. A scene groups temporally and semantically
related events with shared participants, state changes, or development
signals; material bundles may receive a short model-generated summary, while
the underlying events remain authoritative. Plotline candidates require more
than a shared character: they must also exhibit content continuity, temporal
adjacency, a state or transition signal, or an existing development link.
Plots group connected plotlines and do not force isolated threads into a
larger arc. If a consolidation attempt fails its checks, the lower-level
records are retained and no unsupported summary is committed.

Every scene, plotline, and plot stores its member records, source units,
supporting evidence, temporal range, participants, and uncertainty. The
resulting atomic-closure operation expands any selected high-level node back
to its constituent events, propositions, and exact source spans. Retrieval can
therefore use a compact summary as an entry point without losing the evidence
needed to inspect or regenerate the underlying narrative state.

\subsection{Worked multi-view update}
\label{app:construction-example}

Consider the source unit ``Mara moved the key from the drawer into the hollow
book while Leo was away.'' Entity grounding links Mara, Leo, the key, and the
book to existing records. Event extraction adds a movement event with the
location transition as its effect. The world projection records that the key
is now in the book and closes the earlier drawer-location interval. Mara's
belief view receives the new location through her participation, whereas Leo's
belief view remains unchanged because the unit contains no observation,
communication, or inference linking him to the move. The event advances the
development in which Leo is searching for the key. If that development remains
open, the possibility view may retain conditional candidates such as Mara
informing Leo or Leo discovering the key; these candidates are marked as
hypothetical and are never written back as established world facts. This
single update illustrates why the four views share entities and events while
preserving different evidential and epistemic status.

\subsection{Selected Prompt Instructions}
\label{app:selected-prompts}

The following abridged instructions summarize memory formation and
writing-time planning. Illustrative examples clarify the distinctions among
events, facts, beliefs, developments, and possible worlds.

\paragraph{Narrative Memory Planning.}
\label{app:planning-prompt}

The planner selects relevant evidence and focal characters, then identifies
information still needed for the next writing step.

\begin{swpmprompt}{Narrative Memory Planning (abridged)}
\textbf{Inputs:} the writing request, current passage, admissible memory
preview, complete character roster, and reference examples.

\medskip
\textbf{Task}\par
Select established narrative memory needed for the next writing step.
W contains shared-world facts, events, states, and relationships.
B contains character-specific knowledge, beliefs, doubts, and misunderstandings.
D connects supported changes across events, including progress, reversal,
cause, consequence, and relationship development.
H contains conditional possibilities, their premises, and possible consequences.

\medskip
\textbf{Decision rules}\par
Preserve exact names, quantities, object states, time expressions, knowledge
boundaries, and unresolved commitments. Select related records together when
they explain one developing situation. Keep the underlying details available
alongside summaries.\par
Form a query describing what is needed from memory. Add information needs
only for material not already covered by the selected evidence. Distinguish
facts, temporal and causal relations, motivation, character knowledge, state
history, unresolved threads, developments, and possible worlds.\par
Use only characters and evidence supplied in the admissible context. Briefs
and outlines express intentions, not evidence that planned events occurred.
Do not write the next passage.

\medskip
\textbf{Illustrative response}\par
Selected evidence: Mara's promise and the related world, belief, and
development records. Focal characters: Mara and Leon. Remaining information
need: what Leon knows about the available route at this point.
This example assumes that the selected evidence leaves that question unanswered.
\end{swpmprompt}

\paragraph{Narrative Event and Fact Modeling.}
\label{app:event-prompt}

Event modeling separates consequential narrative beats from the precise facts
that describe them, preserving detail without treating every movement as an event.

\begin{swpmprompt}{Narrative Event and Fact Modeling (abridged)}
\textbf{Inputs:} the current passage, source context, identified entities,
unresolved references, supporting passages, and reference examples.

\medskip
\textbf{Task}\par
Extract narrative events and independently readable propositions. An event
changes world state, character knowledge or commitment, a relationship, an
active conflict, or an unresolved development. A proposition records factual
content with its subject, predicate, object, polarity, and truth status.

\medskip
\textbf{Decision rules}\par
Read the passage as a whole. Merge consecutive actions that constitute one
consequential beat. Ask what remains different after the event. Preserve
distinguishing names, objects, quantities, attributes, dates, and durations
in propositions.\par
Support every event and proposition with a source passage. Earlier summaries
provide continuity context, not evidence for new records. A plan is not its
execution, chronology is not causation, and a spoken claim does not establish
its truth. Retain false or disputed claims when needed to represent a
character's belief. Leave unresolved identities explicit.

\medskip
\textbf{Illustrative example}\par
Passage: ``Mara handed the key to Leon.''\par
Event: Mara transfers the key to Leon; participants are Mara, Leon, and the key.\par
Resulting fact: Leon possesses the key.\par
Both records cite the passage. The transfer is one event, while the resulting
possession remains separately recoverable.
\end{swpmprompt}

\paragraph{Character Knowledge Modeling.}
\label{app:belief-prompt}

Perspective modeling distinguishes world propositions from each character's
access to and stance toward them. Each belief remains linked to its supporting
evidence and acquisition event.

\begin{swpmprompt}{Character Knowledge Modeling (abridged)}
\textbf{Inputs:} the current passage, source context, identified entities,
grounded events and propositions, and reference examples.

\medskip
\textbf{Task}\par
Identify the supported discourse perspective and material changes or differences
in character access to facts. For each belief, retain the holder, proposition,
attitude, acquisition mode, and acquisition event when established.

\medskip
\textbf{Decision rules}\par
Ground access in seeing, hearing, reading, discovery, communication,
remembering, inference, doubt, or explicit lack of information. Being named
in an event does not imply that a character knows every fact associated with it.\par
A speaker's assertion does not prove that the speaker believes it, that a
listener accepts it, or that the claim is true. Preserve speakers, recipients,
and distinct belief holders. A question indicates inquiry or uncertainty,
not necessarily disbelief. Emotion and intention alone do not establish
knowledge.\par
Record unawareness only when the passage explicitly establishes a character's
lack of information. Missing evidence is not evidence of unawareness.

\medskip
\textbf{Illustrative example}\par
Passage: ``The lamp was broken. Iona falsely told Tomas it worked; Tomas
believed her.''\par
World proposition: ``The lamp works'' is false.\par
Character belief: Tomas believes that proposition after hearing it from Iona.\par
The belief is linked to Iona's statement as its acquisition event. The passage
does not establish that Iona believes her own claim.
\end{swpmprompt}

\paragraph{Development and Possible-World Modeling.}
\label{app:development-prompts}

State and thread formation preserves established commitments and their changing
status. Possible worlds are conditional extensions of unresolved developments;
consolidation organizes accepted developments across scenes.

\begin{swpmprompt}{State, Thread, and Possible-World Modeling (abridged)}
\textbf{Inputs:} the current passage, entities, grounded events and facts,
character knowledge, relevant prior state, and reference examples.

\medskip
\textbf{Task}\par
Extract supported relationship and entity-state changes and unresolved narrative
threads. Associate conditional possibilities with newly established or
materially changed unresolved threads.

\medskip
\textbf{Decision rules}\par
Keep goals, intentions, commitments, and observed actions distinct. Compare
current evidence with prior state to identify change. Keep the same obligation
identifiable across updates. Mark it resolved only when its promised outcome
is achieved, abandoned when explicitly relinquished, and advanced when progress
leaves that outcome unsettled. A scene change or missing mention does not close
a thread.\par
Construct possibilities from the unresolved condition at the end of the passage.
Do not repeat completed steps. Each possibility retains a supported premise,
possible events, continuity constraints, and uncertainty. Respect character
knowledge and distinguish different mechanisms or choices from paraphrases.
Resolved or abandoned threads have no possibilities; an unresolved thread may
remain without a warranted candidate continuation.

\medskip
\textbf{Illustrative example}\par
Prior commitment: Mara promises to get Leon out tonight.\par
Current passage: ``Mara finally unlocked the gate. Leon was still inside the
locked infirmary, and she had not yet reached him.''\par
Development: the promise has advanced but remains unfulfilled.\par
Possible continuation: if Mara gains entry to the infirmary, she could bring
Leon outside.\par
Constraints: Leon has not been rescued, and entry to the infirmary remains
uncertain. The conditional continuation is not an accepted event.
\end{swpmprompt}

\paragraph{Scene-to-Plotline Consolidation.}
\label{app:scene-plotline-prompt}

\begin{swpmprompt}{Scene-to-Plotline Consolidation (abridged)}
\textbf{Inputs:} source-grounded scene records and reference examples.

\medskip
\textbf{Task}\par
Determine whether the supplied scenes form a continuing plotline: a persistent
goal, conflict, relationship, plan, or state trajectory. Use accepted scenes,
excluding hypothetical possibilities.

\medskip
\textbf{Decision rules}\par
Shared characters alone do not establish continuity. Connect scenes through
supported actions, changes, temporal order, or explicit relations. Distinct
descriptions of the same events are not successive scenes. Distinguish plans
from execution and chronology from causation. Preserve supported resolution,
revelation, and setup/payoff links without inventing them from genre expectations.\par
Summarize the starting situation, the change, and its result. Retain exact
names, quantities, unresolved conditions, and uncertainty when relevant.
Support each connection with at least two distinct scenes.

\medskip
\textbf{Expected response}\par
A concise account of the development, the relation connecting its scenes,
the supporting scene references, and any unresolved uncertainty.
\end{swpmprompt}

\section{Writing Benchmark Adaptations}
\label{app:writing-adaptations}

This appendix specifies how the three writing benchmarks are presented to
memory systems. They test related but distinct capabilities. TMAS is a
single fixed-prefix continuation problem: the system must recover useful
context from a visible source prefix before producing one continuation.
ConStory-Bench is a task-adaptive continual protocol: the system must preserve
a story state across several accepted writing units while the requirements
changes across four task types. LSGC extends continual generation to 14
chapters and at least 40,000 words per story. All conditions receive the same
task inputs.

Across all three evaluations, reference continuations, future text, and
evaluation annotations are withheld from memory construction and generation.
Each story and condition maintains independent memory. The writer receives the
task, available recent history, and retrieved memory as readable text. Full
Context receives the complete available history under the same task and writing
model, while memory systems retain their native construction and retrieval
procedures and process source text or accepted writing units in order.

\subsection{TMAS: source-faithful rolling checkpoints}
\label{app:tmas-adaptation}

The released source contains 168 checkpoints: 120 from the development split
and 48 from the held-out split. We evaluate all 168 checkpoints using a paired
rolling protocol. Each checkpoint contains an English writing instruction, the
source story up to a designated boundary, and an evaluator-only human
continuation. The task instruction is retained verbatim, including its
requested setting, characters, constraints, style, form, and any example-specific
whole-story length instruction. The visible source history is represented as
ordered lines or paragraphs with source positions. The boundary is non-empty
and is selected from the source story rather than from the human continuation.

Each checkpoint is evaluated as follows:

\begin{enumerate}
    \item initialize independent memory for the checkpoint;
    \item ingest the visible prefix in source order, without reading the
    evaluator-only continuation;
    \item issue the memory request at the continuation boundary; and
    \item give the writer the original instruction, the available recent
    prefix, and the memory system's rendered context.
\end{enumerate}

The writer returns only the new prose continuation and must not repeat the
visible prefix. The human continuation is reserved for evaluation. We measure
generated word count after completion, with length requirements inherited from
each original instruction rather than imposed as a universal target. TMAS
therefore evaluates one continuation from a fixed source history.

Each memory system constructs and retrieves context from the visible prefix;
the Full Context control receives the same task and full visible source prefix
directly, without memory construction or retrieval. The reported primary TMAS
dimensions are Plot, Creativity, Development, and Language Use. Table~\ref{tab:tmas-auxiliary}
also reports the arithmetic-mean aggregate, TMAS Quality, together with
Prefix Continuity, generated word counts, and Unique.

\begin{table}[t]
\caption{TMAS aggregate and auxiliary diagnostics. TMAS Quality is the
arithmetic mean of the four reported dimensions; Prefix Continuity, word
counts, and Unique are descriptive measures.}
\label{tab:tmas-auxiliary}
\centering
\small
\setlength{\tabcolsep}{5pt}
\begin{tabular}{lrrrr}
\toprule
Method & TMAS Quality & \shortstack{Prefix\\Continuity $\uparrow$} & \shortstack{Words\\Generated} & Unique \\
\midrule
NarraWorld & \textbf{83.01} & 87.29 & 905.6 & 0.3372 \\
Mem0 & 81.79 & 86.98 & 931.6 & 0.3582 \\
Graphiti & 82.04 & 85.83 & 914.7 & 0.3599 \\
Hindsight & 77.87 & 80.48 & 920.9 & 0.3573 \\
AriGraph & 76.78 & 77.88 & 901.9 & 0.3595 \\
StructMem & 81.80 & 86.83 & 936.8 & 0.3414 \\
Scaffold Only & 81.05 & 68.50 & 922.3 & 0.3532 \\
Full Context & 82.87 & 88.23 & 913.4 & 0.3621 \\
\bottomrule
\end{tabular}
\end{table}

\subsection{ConStory-Bench: task-adaptive continual writing}
\label{app:constory-adaptation}

The evaluation dataset contains 200 English stories, with 50
stories for each of four task types: generation, continuation, expansion, and
completion. Task construction uses only the original prompts, excluding
reference stories and evaluation annotations. DeepSeek-V4-Pro restructures
each prompt into a task-specific global context. Writing schedules, local
objectives, and word targets are fixed by task type and shared across conditions.

The four task types use distinct writing schedules:

\begin{itemize}
    \item \textbf{Generation.} The writer starts from a premise and a set of
    fixed narrative requirements, together with style and form requirements.
    The story is developed over 8 units, targeting approximately 1,100 words
    per unit and 8,800 words in total.
    \item \textbf{Continuation.} The writer receives a supplied opening
    fragment, an explicit continuation boundary, and requirements inherited
    from the original prompt. New prose begins after the opening and must not
    repeat it. This task uses 5 units of approximately 1,760 words, again
    targeting 8,800 words in total.
    \item \textbf{Expansion.} The writer receives an ordered compact outline
    whose beats define the material to be expanded, together with expansion,
    style, and form requirements. The outline is a planning input, not a
    reference story. The task uses 6 units of approximately 1,467 words.
    \item \textbf{Completion.} The writer receives a beginning anchor and an
    ending anchor and must construct the missing bridge while satisfying the
    bridge requirements. The two anchors constrain the endpoint of the story,
    so this is not equivalent to ordinary forward continuation. It uses 6 units
    of approximately 1,467 words.
\end{itemize}

The writer receives task-specific instructions: a premise, an opening and
continuation boundary, an ordered outline, or beginning and ending anchors.
Across all four types, the dataset contains 1,250
writing units. The first unit of each story establishes the initial
writing state, so 1,050 later units can retrieve material
from previously accepted writing.

For each story, the continual protocol is:

\begin{enumerate}
    \item initialize independent story memory and expose the task's global
    context and first unit's requirements;
    \item generate and accept the next narrative unit;
    \item ingest the accepted unit into the memory system before the next
    unit; and
    \item retrieve a new memory view for the next unit using the global task
    requirements, the current unit goal, and the available generated history.
\end{enumerate}

The current unit's instructions state its local goal and any required material;
the global context preserves the original requirements and the task-specific
anchors or outline. The output is narrative prose with the
unit-level word target. The final story is the ordered concatenation of the
accepted units. Future units, evaluator targets, and other conditions are not
visible when an earlier unit is generated.

This protocol makes ConStory a test of memory under changing writing states.
The memory system must retain characters, events, world facts, private
knowledge, and unresolved developments that remain useful later, while the
writer must also respect the current unit's requirements. The four task
types differ in the initial state and in the constraints that must remain
available; they are therefore reported separately as well as in aggregate.

The official ConStory evaluator is applied after all units are assembled.
We report its Consistency Error Density (CED, lower is better) and the official
category-level consistency breakdown.

\subsection{LSGC: chapter-continual long-story generation}
\label{app:lsgc-adaptation}

The LSGC memory track uses all five released official starters. Each system
develops every starter into a 14-chapter story, with two sequential generation
segments per chapter, and repeats the complete protocol for three runs. The
target is 42,000 words per story, and a completed story must contain at least
40,000 words. All starters and runs use the same chapter-by-chapter writing
protocol.

Before generation, the writer receives the starter and a reviewed
coarse-grained outline shared by all memory conditions. For chapter $t$, the
legal input consists of the starter, the same outline, the current chapter
objective, the complete text of chapter $t-1$, and the memory system's rendered
view of chapters earlier than $t$. The memory ceiling is 12,000 rendered tokens
and does not include the shared starter, outline, chapter objective, or complete
preceding chapter. Memory is retrieved once at the start of a chapter and reused
for its two segments. After both segments are accepted, the new chapter is
incorporated into memory before the next chapter.

We retain the Longform Writing Bench aggregation structure while adapting the
final-piece context to the much longer LSGC stories. A condition-blind Judge
scores the 16 criteria for each chapter. After the fourteenth chapter is
accepted, the system produces a final rolling summary over the complete story,
and the Judge scores the final piece from the writing prompt, starter, and this
summary rather than the full 40,000-word story. We then average the
14 chapter judgments and the final-piece judgment within each story instance.
Raw criteria retain their original 0--20 scales and directions:
positive criteria are multiplied by five, while negative criteria are mapped
to $(20-s)\times5$.

For compact reporting, Character averages Believable Character Actions,
Nuanced Characters, and Unearned Transformations. Plot averages Pacing,
Compelling Plot, Emotionally Engaging, and Well-earned Lightness or Darkness.
World/Coherence averages World Building and Coherent. Style/Creativity averages
Weak Dialogue, Tell-Don't-Show, Unsurprising or Uncreative, Amateurish, Purple
Prose, and Forced Poetry or Metaphor. Prompt Adherence contains Faithful to
Writing Prompt. Overall is the equal-weight mean of all 16 criteria, not an
equal-weight mean of the five groups. The completed story, after this
within-story aggregation, is the primary statistical unit; chapter-level
judgments are not reported separately or treated as independent samples.

\begin{table*}[t]
\caption{LSGC final-rubric scores by criterion. Values are direction-corrected
macro means over five starters and three runs on a 0--100 scale ($\uparrow$);
negative criteria are inverted.}
\label{tab:lsgc-criteria}
\centering
\footnotesize
\setlength{\tabcolsep}{1pt}
\begin{tabular}{lrrrrrrrr}
\toprule
Criterion & Mem0 & AriGraph & Hindsight & Graphiti & StructMem & \shortstack{Narra\\World} & \shortstack{Scaffold\\Only} & \shortstack{Full\\Context} \\
\midrule
Believable Character Actions & 89.07 & 78.13 & 80.84 & 88.69 & 90.02 & \textbf{91.04} & 70.96 & 86.80 \\
Nuanced Characters & 87.62 & 72.27 & 75.78 & 88.51 & 88.22 & 91.22 & 69.11 & \textbf{91.53} \\
Pacing & 70.93 & 71.02 & 72.98 & 81.29 & 85.20 & \textbf{87.71} & 69.09 & 81.76 \\
World Building & 86.00 & 74.80 & 79.06 & 88.29 & 87.29 & 87.49 & 66.56 & \textbf{92.62} \\
Compelling Plot & 80.22 & 75.62 & 79.91 & \textbf{88.76} & 85.60 & 87.25 & 67.18 & 88.42 \\
Emotionally Engaging & 87.58 & 80.53 & 82.38 & 89.00 & 89.84 & \textbf{94.60} & 71.02 & 91.09 \\
Coherent & 87.56 & 78.22 & 81.49 & 88.93 & 91.11 & \textbf{92.02} & 66.18 & 91.13 \\
Weak Dialogue & 74.73 & 72.00 & 72.64 & 73.18 & 85.60 & \textbf{91.00} & 72.20 & 77.56 \\
Tell-Don't-Show & 65.16 & 69.24 & 66.78 & 67.58 & 80.09 & \textbf{86.87} & 67.80 & 74.51 \\
Unsurprising or Uncreative & 77.33 & 71.60 & 73.07 & 60.82 & 78.18 & \textbf{84.60} & 65.76 & 77.53 \\
Amateurish & 77.60 & 76.02 & 76.29 & 71.89 & 87.98 & \textbf{93.16} & 74.80 & 87.16 \\
Purple Prose & 74.00 & 55.98 & 56.33 & 63.44 & 85.16 & \textbf{89.67} & 71.73 & 78.44 \\
Forced Poetry or Metaphor & 73.29 & 60.53 & 61.27 & 67.13 & 84.62 & \textbf{88.56} & 72.87 & 77.56 \\
Unearned Transformations & 76.33 & 67.27 & 69.33 & 70.11 & 88.38 & \textbf{91.09} & 59.42 & 83.18 \\
Well-earned Lightness or Darkness & 88.09 & 80.31 & 82.58 & 89.40 & 90.31 & \textbf{92.42} & 69.60 & 87.82 \\
Faithful to Writing Prompt & 87.40 & 85.00 & 87.87 & 88.51 & 93.07 & \textbf{95.09} & 79.53 & 93.42 \\
\bottomrule
\end{tabular}
\end{table*}

\begin{table}[t]
\caption{LSGC Overall scores by starter, averaged over three runs. Each value
equally weights the 16 direction-corrected final criteria.}
\label{tab:lsgc-starters}
\centering
\small
\setlength{\tabcolsep}{1.5pt}
\begin{tabular}{lrrrrrrrr}
\toprule
Starter & Mem0 & AriGraph & Hindsight & Graphiti & StructMem & NarraWorld & \shortstack{Scaffold\\Only} & \shortstack{Full\\Context} \\
\midrule
01 & 79.33 & 73.32 & 74.17 & 85.78 & 84.83 & \textbf{91.95} & 74.55 & 87.26 \\
02 & 65.67 & 73.07 & 75.60 & 85.36 & \textbf{89.35} & 86.78 & 61.74 & 86.35 \\
03 & 85.47 & 71.56 & 73.36 & 54.53 & 86.61 & \textbf{91.05} & 65.81 & 88.03 \\
04 & 82.93 & 74.26 & 77.20 & 84.69 & 87.52 & \textbf{90.38} & 84.13 & 87.48 \\
05 & 87.51 & 72.97 & 74.24 & 85.11 & 86.27 & \textbf{91.01} & 61.85 & 76.04 \\
\midrule
Macro mean & 80.18 & 73.03 & 74.91 & 79.10 & 86.92 & \textbf{90.24} & 69.61 & 85.03 \\
\bottomrule
\end{tabular}
\end{table}

\begin{table*}[t]
\caption{Run-to-run variability of LSGC Overall scores. Each run is first
aggregated over five starters and the 16 criteria; range and SD are computed
over the three independent runs. SD is the sample standard deviation.}
\label{tab:lsgc-run-variability}
\centering
\small
\setlength{\tabcolsep}{5pt}
\begin{tabular}{lrrr}
\toprule
Method & Overall mean $\uparrow$ & Run range & Run SD \\
\midrule
NarraWorld & \textbf{90.24} & 2.59 & 1.35 \\
Mem0 & 80.18 & 5.38 & 2.69 \\
Graphiti & 79.10 & 3.03 & 1.51 \\
Hindsight & 74.91 & 4.06 & 2.10 \\
AriGraph & 73.03 & 3.57 & 1.90 \\
StructMem & 86.92 & 3.34 & 1.69 \\
Scaffold Only & 69.61 & 4.55 & 2.30 \\
Full Context & 85.03 & 2.13 & 1.10 \\
\bottomrule
\end{tabular}
\end{table*}

Let $y_{m,r}$ denote the direction-corrected Overall score
for method $m$ in run $r$, after averaging that run over the five starters and
the 16 criteria. With $R=3$ runs, the reported mean, range, and sample standard
deviation are respectively
\begin{equation}
\bar y_m=\frac{1}{R}\sum_{r=1}^{R}y_{m,r},\qquad
\operatorname{Range}_m=\max_r y_{m,r}-\min_r y_{m,r},
\end{equation}
\begin{equation}
s_m=\sqrt{\frac{1}{R-1}\sum_{r=1}^{R}(y_{m,r}-\bar y_m)^2}.
\end{equation}
Range and SD describe variation among the three complete runs after within-run
aggregation; they are not confidence intervals.

% Appendix tables are defined here so that all manuscript tables remain in one file.
\newcommand{\constoryErrorTaxonomyTable}{%
\begin{table}[htbp]
\caption{ConStory-Checker's five categories and 19 error subtypes.}
\label{tab:constory-error-taxonomy}
\centering\small
\begin{tabular}{p{0.22\linewidth}p{0.71\linewidth}}
\toprule
Category & Subtypes \\
\midrule
Characterization & Memory contradictions; knowledge contradictions;
skill/power fluctuations; forgotten abilities. \\
Factual detail & Appearance mismatches; nomenclature confusion;
quantitative mismatches. \\
Narrative style & Perspective confusion; tone inconsistencies; style shifts. \\
Timeline/plot & Absolute-time contradictions; duration/timeline
contradictions; simultaneity contradictions; causeless effects;
causal-logic violations; abandoned plot elements. \\
World building & Core-rule violations; social-norm violations;
geographical contradictions. \\
\bottomrule
\end{tabular}
\end{table}}

\newcommand{\tmasSubcriteriaTable}{%
\begin{table}[htbp]
\caption{TMAS evaluation subcriteria, each scored from 0 to 4.}
\label{tab:tmas-subcriteria}
\centering\small
\begin{tabular}{p{0.20\linewidth}p{0.73\linewidth}}
\toprule
Dimension & Subcriteria \\
\midrule
Plot & Connected structure; plot progression; logical coherence;
intentional turns. \\
Creativity & Engaging specificity; non-generic ideas; purposeful use of
tropes; original contribution. \\
Development & Character context; setting context; believable detail;
motivated development. \\
Language Use & Sentence variety; lexical richness; literary effect;
repetition control; appropriateness of voice. \\
Prefix Continuity & Continuity of events, characters, perspective and voice,
and facts. \\
\bottomrule
\end{tabular}
\end{table}}

\newcommand{\rmtbenchDimensionsTable}{%
\begin{table}[htbp]
\caption{RMTBench dimensions used by the role-play Judge.}
\label{tab:rmtbench-dimensions}
\centering\small
\begin{tabular}{lp{0.81\linewidth}}
\toprule
Dimension & Interpretation \\
\midrule
EE & Emotional Expression: natural, proportionate emotion appropriate to the situation. \\
EC & Emotional Comprehension: recognizes and responds appropriately to explicit and implicit user emotions. \\
PA & Plot Advancement: enriches the conversation with relevant information, discussion points, or scenarios. \\
CU & Character Understanding: correct identity, background, relationships, and situation. \\
CM & Character Maintenance: stable role and voice without identity drift or generic assistant framing. \\
UPA & User Preference Awareness: identifies and applies implicit or explicit user preferences across turns. \\
\bottomrule
\end{tabular}
\end{table}}

\newcommand{\locomoBudgetResultsTable}{%
\begin{table*}[t]
\caption{LoCoMo budget sensitivity for NarraWorld, StructMem, and Mem0. All scores
are higher-is-better.}
\label{tab:locomo-budget-results}
\centering
\footnotesize
\setlength{\tabcolsep}{2pt}
\begin{tabular*}{\textwidth}{@{\extracolsep{\fill}}llrrrrr@{}}
\toprule
Method & Budget & \shortstack{Overall\\Main} & Multi-hop & Single-hop & Temporal & Open-domain \\
\midrule
Mem0 & 2k & 0.4736 & 0.3718 & 0.5241 & 0.4919 & 0.2691 \\
     & 4k & 0.4982 & 0.4047 & 0.5392 & 0.5372 & \textbf{0.2834} \\
     & 8k & \textbf{0.5010} & \textbf{0.4100} & \textbf{0.5410} & \textbf{0.5470} & 0.2700 \\
     & 12k & 0.4683 & 0.3984 & 0.5125 & 0.4758 & 0.2620 \\
     & 16k & 0.4742 & 0.4047 & 0.5125 & 0.4963 & 0.2682 \\
\midrule
StructMem & 2k & 0.4962 & 0.3565 & 0.5275 & 0.5827 & 0.3435 \\
          & 4k & 0.5324 & 0.4109 & 0.5659 & \textbf{0.6020} & 0.3631 \\
          & 8k & 0.5400 & 0.4350 & 0.5750 & 0.5950 & 0.3620 \\
          & 12k & 0.5365 & 0.4311 & 0.5725 & 0.5897 & 0.3526 \\
          & 16k & \textbf{0.5442} & \textbf{0.4352} & \textbf{0.5836} & 0.5892 & \textbf{0.3680} \\
\midrule
NarraWorld & 2k & 0.5605 & 0.4225 & 0.5985 & 0.6405 & 0.3655 \\
     & 4k & 0.5878 & 0.4386 & 0.6310 & 0.6567 & 0.4173 \\
     & 8k & 0.6144 & 0.4580 & 0.6790 & 0.6490 & 0.3920 \\
     & 12k & 0.6211 & 0.4499 & 0.6853 & 0.6515 & \textbf{0.4594} \\
     & 16k & \textbf{0.6238} & \textbf{0.4712} & \textbf{0.6897} & \textbf{0.6577} & 0.3810 \\
\bottomrule
\end{tabular*}
\end{table*}}

\section{Evaluation Protocols, Metrics, and Judge Prompts}
\label{app:evaluation-protocols}

This appendix specifies the metric sources, scoring procedures, and
task-specific adaptations. Judges evaluate anonymized outputs against the
task evidence; reference answers and evaluation annotations are reserved for
scoring.

\subsection{ConStory-Bench: consistency error density}
\label{app:constory-metrics}

We use the official ConStory-Bench taxonomy and checker
\citep{li2026constory}. The five categories cover 19 error subtypes, and CED
counts whether each subtype has at least one supported error in the assembled
story. For story $i$, let $L_i$ be its whitespace-delimited word count and
$z_{ic}$ indicate the presence of subtype $c$:
\begin{equation}
 E_i=\sum_{c=1}^{19} z_{ic},\qquad
 \mathrm{CED}_i=10{,}000\frac{E_i}{L_i},\qquad
 \mathrm{CED}=\frac{1}{N}\sum_{i=1}^{N}\mathrm{CED}_i.
\end{equation}
Multiple instances of one subtype therefore contribute one unit. Task-specific
columns use the same story-level mean within each task type.

Words per Penalized Error (WPE) is an auxiliary measure introduced for this
paper, adapted from the benchmark's length-normalized intermediate score:
\begin{equation}
 \mathrm{WPE}=\frac{1}{N}\sum_{i=1}^{N}\frac{L_i}{1+E_i}.
\end{equation}
Both measures depend on story length and are reported as consistency
diagnostics rather than literary-quality scores.

\subsection{Tell Me A Story: adapted continuation quality}
\label{app:tmas-metrics}

The original Tell Me A Story evaluation uses pairwise comparisons and a
Bradley--Terry aggregation. We use its Plot, Creativity, Development, and
Language Use dimensions for the fixed-prefix continuation task.
Each Judge request contains the original writing prompt, the shared prefix,
the candidate continuation, and the evaluator-only human continuation.
The reference calibrates execution quality; it is not the only acceptable
plot or ending.

For checkpoint $i$, dimension $d$ has $K_d$ integer subscores
$a_{idk}\in\{0,1,2,3,4\}$. Plot, Creativity, and Development each have four
subcriteria; Language Use has five. We normalize each dimension and define our
arithmetic-mean aggregate as
\begin{equation}
 S_{id}=\frac{100}{4K_d}\sum_{k=1}^{K_d}a_{idk},\qquad
 Q_i=\frac{1}{4}\sum_{d\in\{\mathrm{Plot,Creativity,Development,Language}\}}S_{id}.
\end{equation}
Reported dimension scores and TMaS Quality are arithmetic means over completed
checkpoints. The four dimensions receive equal weight; Language's fifth
subcriterion does not give Language extra weight in $Q_i$.

\textbf{Prefix Continuity} is an auxiliary measure introduced for our
continuation task, not a source-benchmark metric. Its four subscores measure
event, character/motivation, perspective/voice, and factual continuity with
the supplied prefix. It does not enter $Q_i$. Generated Words counts only new
continuation text, and Unique is the distinct-word ratio; both are descriptive
diagnostics.

\begin{swpmprompt}{Tell Me A Story Judge: scoring instructions}
Evaluate one continuation in the context of the original writing prompt and
shared prefix. Use the human continuation only as a fixed quality and
task-interpretation anchor. Do not select a winner or compare system
identities. Score every named subcriterion with one integer from 0 to 4,
where 0 indicates failure, 2 competent execution with a clear limitation, and
4 exceptional execution without a meaningful weakness. Support scores with
brief grounded evidence and also evaluate event, character, perspective, and
factual continuity with the prefix.

\textbf{Inputs:} the original writing prompt, shared story prefix, candidate
continuation, and evaluator-only human continuation.
\end{swpmprompt}

\tmasSubcriteriaTable

\subsection{LSGC: full-story quality under the Longform Writing Bench rubric}
\label{app:lsgc-metrics}

LSGC specifies a long-form generation protocol but does not prescribe a native
quality metric \citep{migal2024lsgc}. We therefore evaluate its completed stories with the separate
16-criterion rubric from Longform Writing Bench. We retain the criterion names
and raw 0--20 scores, and report direction-corrected 0--100 scores in five
groups alongside an equal-weight Overall.

For story instance $i$, let $b_{ijc}\in[0,20]$ be the raw score for criterion
$c$ at chapter or final-piece position $j$, where
$j\in\{1,\ldots,14,\mathrm{final}\}$. For the seven negative
criteria---Weak Dialogue, Tell-Don't-Show, Unsurprising or Uncreative,
Amateurish, Purple Prose, Forced Poetry or Metaphor, and Unearned
Transformations---lower raw values indicate fewer problems. We first apply the
direction correction to every chapter and final-piece score and then average
the resulting 15 values within each story instance:
\begin{equation}
 T_{ijc}=\begin{cases}
 5(20-b_{ijc}), & c\text{ is negative},\\
 5b_{ijc}, & c\text{ is positive},
 \end{cases}
 \qquad
 \bar T_{ic}=\frac{1}{15}\sum_{j=1}^{15}T_{ijc},
 \qquad O_i=\frac{1}{16}\sum_{c=1}^{16}\bar T_{ic}.
\end{equation}
Each reported column is then a macro mean over five starters and three runs. A
grouped column averages its assigned criteria within each story instance
before taking the same macro mean. The reported unit is therefore the
completed story instance, not an individual chapter.
Character, Plot, World/Coherence, Style/Creativity, and Prompt Adherence use
the group membership specified in Appendix~\ref{app:lsgc-adaptation}; Overall
averages all 16 criteria rather than the five group scores.

For run-level variability, let $u_{mrc}$ be the mean score for method $m$,
run $r$, and criterion $c$ after averaging the five starters, and let
$o_{mr}=16^{-1}\sum_{c=1}^{16}u_{mrc}$ be that run's Overall score. Across the
three runs, the reported mean, range, and sample standard deviation are
\begin{equation}
\mu_m=\frac{1}{3}\sum_{r=1}^{3}o_{mr},\qquad
R_m=\max_r o_{mr}-\min_r o_{mr},\qquad
s_m=\sqrt{\frac{1}{2}\sum_{r=1}^{3}(o_{mr}-\mu_m)^2}.
\label{eq:lsgc-run-variability}
\end{equation}
The criterion-level range and standard deviation use the same formulas with
$o_{mr}$ replaced by $u_{mrc}$. Thus the reported Overall variability is
computed after averaging criteria, whereas criterion-level variability retains
variation specific to each rubric item.

\begin{swpmprompt}{LSGC Judge (core instructions)}
Evaluate the current chapter or the final story-level view against the supplied
writing prompt and all 16 criteria. For the final-piece judgment, use the
official starter and the final rolling summary produced after Chapter 14.
Score every criterion from 0 to
20 and briefly justify the scores with concrete strengths and weaknesses. For
the seven negative criteria listed above, lower scores indicate fewer problems.

\textbf{Inputs:} the chapter text for chapter judgments, or the official
starter and final rolling summary for the final-piece judgment, together with
the 16 Longform Writing Bench criteria.
\end{swpmprompt}

\subsection{RMTBench: turn-level role-play scores}
\label{app:rmtbench-metrics}

We follow RMTBench's native mixed-scale evaluation
\citep{xiang2025rmtbench}. EE, EC, PA, and CU use integer scores from 1 to 5;
CM and UPA use binary success values. The four Likert dimensions apply to all
scenarios, CM applies to the character-maintenance scenario, and UPA applies
to the final preference-recognition turns.

For metric $d$, let $\mathcal A_d$ be its applicable turns and $r_{td}$ its
value. We use the turn-level mean
\begin{equation}
 \bar R_d=\frac{1}{|\mathcal A_d|}\sum_{t\in\mathcal A_d}r_{td}.
\end{equation}
Likert dimensions remain on their 1--5 scale, while binary dimensions are
reported as $100\bar R_d\%$. Inapplicable dimensions are excluded from their
own denominator, and no cross-metric average is reported.

\begin{swpmprompt}{RMTBench Role-Play Judge (core instructions)}
Evaluate only the dimensions supplied as applicable. Use the original 1--5
rubrics for EE, EC, PA, and CU. Score CM and UPA as binary success or failure
according to their applicable RMTBench criteria, and report null for every
inapplicable dimension.
\end{swpmprompt}

\subsection{LoCoMo: category-aware deterministic scoring}
\label{app:locomo-metrics}

We use the released category-aware LoCoMo answer scorer
\citep{maharana2024locomo}. The main comparison evaluates the C1--C4 subset:
282 multi-hop, 321 temporal, 96 open-domain, and 841 single-hop questions,
for 1,540 questions in total. The 446 adversarial C5 questions are excluded.
Each category score follows the official scorer, while Overall Main is the
question-weighted mean over these 1,540 main questions.

\locomoBudgetResultsTable

\paragraph{Comparison controls and length measures.}
\label{app:evaluation-integrity}

Comparisons use the same task set and evaluation protocol across conditions.
Scores include only completed generations with valid assessments.
Each condition develops its own
history throughout generation. ConStory and LSGC word counts use
whitespace-delimited English words. TMaS counts only newly generated
continuation words. Memory length is measured in tokens and excludes the
writing instruction, recent legal text supplied outside memory, and generated
output. Retrieved memory may occupy less than the specified ceiling.

\section{Retrieval Planning Ablations}
\label{app:retrieval-ablation}

\begin{figure}[!htbp]
\centering
\includegraphics[width=\textwidth]{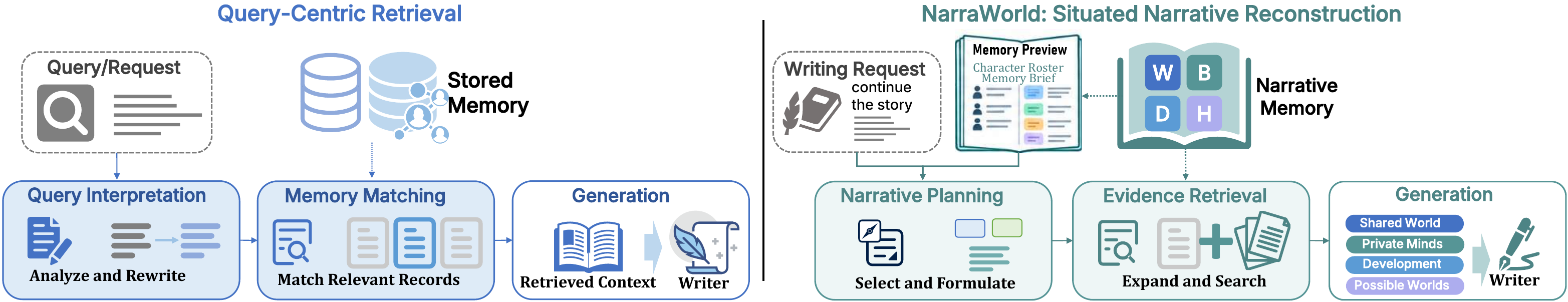}
\caption{Conceptual comparison of retrieval paradigms. Query-centric methods
interpret the request and match the resulting query against stored memory;
both deterministic parsing and question-only LLM planning instantiate this
left-hand path. NarraWorld instead conditions narrative planning on an admissible
memory preview before expanding and searching for supporting evidence. The
diagram summarizes the two information flows.}
\label{fig:retrieval-comparison}
\end{figure}

We isolate whether retrieval planning benefits from access to a memory preview
on both LoCoMo and ConStory-Bench. Figure~\ref{fig:retrieval-comparison} places
no-preview controls within the query-centric paradigm and contrasts them with
NarraWorld's preview-conditioned reconstruction. The generic diagram depicts the full
NarraWorld memory interface; the benchmark-specific experiments below hold their
respective offline memories, context budgets, and downstream models fixed.

\paragraph{Deterministic Retrieval.}
Rules parse surface entities, temporal expressions, relation terms, and question
forms. Temporal questions activate time and event channels, for example, while
causal questions additionally activate causal and development access. This
setting requires no model call for planning and is fully deterministic, but it
cannot reinterpret implicit intent beyond the encoded rules. Because these
rules map directly to LoCoMo question forms, we use this setting only as a
LoCoMo control rather than as a general writing baseline.

\paragraph{Retrieval w/o Dependency Planning.}
The planner receives the question but no memory content. It extracts target
entities, assigns $W/B/D$ channels, rewrites the query, and specifies temporal,
perspective, or multi-hop constraints. This setting tests semantic question
interpretation without dependency planning grounded in records that actually
exist in memory.

\paragraph{Preview-Guided Reconstruction (NarraWorld).}
Before formal retrieval, the planner receives a lightweight admissible preview
of the $W/B/D$ directory. It may select records that already address the
question or use visible events and entities as anchors for retrieving missing
links. Preview-Guided Reconstruction therefore grounds dependency planning in the available
memory structure before downstream answering.

\subsection{LoCoMo: Question-Centric Retrieval}

All three LoCoMo settings use the same offline $W/B/D$ memory, 8k final memory
budget, and answer-generation model. They differ only in how the question is
converted into retrieval operations and whether the planner can inspect an
admissible summary of stored memory before formal retrieval.

\begin{table}[!htbp]
\caption{LoCoMo retrieval ablation with shared offline $W/B/D$
memory, an 8k final memory budget, and the same answer model. All metrics are
higher-is-better.}
\label{tab:retrieval-planning-ablation}
\centering
\small
\setlength{\tabcolsep}{3.5pt}
\begin{tabular}{lrrrrr}
\toprule
Retrieval setting & \shortstack{Overall\\Main} & Multi-hop & Single-hop & Temporal & Open-domain \\
\midrule
Deterministic Retrieval & 0.5963 & 0.4479 & 0.6467 & \textbf{0.6513} & 0.4069 \\
Retrieval w/o Dependency Planning & 0.5982 & 0.4487 & 0.6488 & 0.6449 & \textbf{0.4383} \\
Preview-Guided Reconstruction (NarraWorld) & \textbf{0.6144} & \textbf{0.4580} & \textbf{0.6790} & 0.6490 & 0.3920 \\
\bottomrule
\end{tabular}
\end{table}

Table~\ref{tab:retrieval-planning-ablation} shows that Preview-Guided Reconstruction
achieved the highest Overall Main score (0.6144), improving over Deterministic
Retrieval by 0.0181 and Retrieval w/o Dependency Planning by 0.0162.
It also led multi-hop and single-hop accuracy. Relative to the LLM query-centric
control, the gains were 0.0093 on multi-hop and 0.0302 on single-hop, consistent
with the benefit of anchoring retrieval decisions in the available memory
structure rather than deriving them from the question alone.

Replacing Deterministic Retrieval with Retrieval w/o Dependency Planning changed Overall
Main by only 0.0019 under the matched protocol. The two query-centric variants
nevertheless showed different category strengths: deterministic rules obtained
the highest temporal score (0.6513), whereas LLM planning obtained the highest
open-domain score (0.4383). Preview-Guided Reconstruction combined the strongest
aggregate result with the best multi-hop and single-hop scores.

\subsection{ConStory-Bench: Writing-Conditioned Retrieval}

Writing does not provide the lexical question forms required by deterministic
retrieval: a continuation request rarely states which prior facts or
developments should be recovered. We therefore compare retrieval without
dependency planning against the full NarraWorld retrieval pipeline. Both receive the
same global plan, rolling summary, recent text, and current writing objective.
The first setting retrieves directly from this scaffold, whereas Preview-Guided Reconstruction
uses the admissible $W/B/D/H$ preview to plan dependencies before retrieval.
The offline memory, retrieved-memory budget, writing model, and generation
protocol are otherwise held fixed.

\begin{table}[!htbp]
\caption{ConStory-Bench retrieval ablation under matched writing
scaffolds, offline memory, memory budgets, and generation settings. All columns
report CED ($\downarrow$).}
\label{tab:constory-retrieval-planning}
\centering
\small
\setlength{\tabcolsep}{3.5pt}
\begin{tabular}{lrrrrr}
\toprule
Retrieval setting & CED $\downarrow$ & Generation & Continuation & Expansion & Completion \\
\midrule
Retrieval w/o Dependency Planning & 0.315 & 0.210 & 0.380 & 0.185 & 0.485 \\
Preview-Guided Reconstruction (NarraWorld) & \textbf{0.185} & \textbf{0.120} & \textbf{0.235} & \textbf{0.098} & \textbf{0.287} \\
\bottomrule
\end{tabular}
\end{table}

Preview-Guided Reconstruction reduced overall CED by 0.130 and improved every
task type. The largest absolute reduction occurred on Completion (0.198),
followed by Continuation (0.145), where selecting earlier constraints and
developments is especially important. These results isolate dependency
planning over the memory preview rather than the contribution of the writing
scaffold itself, which is shared by both conditions.

\section{Memory Ablations Across Tasks}
\label{app:memory-ablations}

The ablation study uses the same four configurations on ConStory-Bench and
TMAS. Full NarraWorld is compared with removal of character beliefs,
possible worlds, or cross-event organization.
The belief ablation excludes character-specific epistemic assertions from
both planner previews and final memory, including assertions carried by
derived summaries. The possible-world ablation disables $H$ formation and
access while retaining unresolved accepted developments in $D$. The
cross-event ablation disables derived scene, plotline, and plot access while
preserving atomic events, world facts, beliefs, and development records.
Models, writing instructions, decoding settings, memory caps, and evaluation
protocols are held fixed across conditions. Contexts are not padded to fill the
cap. TMAS uses the same fixed source histories with
the corresponding memory views excluded. ConStory conditions generate
independent complete trajectories and incorporate their own accepted units;
later memory states from Full NarraWorld are not reused for another condition.
Comparisons are paired by story or checkpoint.

\paragraph{ConStory-Bench consistency and task coverage.}

Figure~\ref{fig:constory-ablation} reports Overall CED and the four task-specific
scores for each configuration.

\subsection{TMAS: continuation quality}

The matched continuation evaluation complements consistency measurement:
possible-world memory may affect creativity and development even when it
does not change consistency error density. Table~\ref{tab:tmas-ablations}
reports the four quality dimensions, their equal-weight TMAS
Quality aggregate, and the auxiliary Prefix Continuity diagnostic.

\begin{table}[htbp]
\caption{TMAS ablations. All quality dimensions are higher-is-better.}
\label{tab:tmas-ablations}
\centering
\small
\setlength{\tabcolsep}{2.5pt}
\begin{tabular}{@{}lrrrrrr@{}}
\toprule
Setting & Plot & Creativity & Development & Language & \shortstack{TMAS\\Quality} & \shortstack{Prefix\\Continuity} \\
\midrule
Full NarraWorld & \textbf{78.92} & \textbf{84.90} & \textbf{85.60} & \textbf{82.62} & \textbf{83.01} & \textbf{87.29} \\
Without character beliefs ($B$) & 76.54 & 82.15 & 81.33 & 81.58 & 80.40 & 83.21 \\
Without possible worlds ($H$) & 77.20 & 78.85 & 80.15 & 82.10 & 79.57 & 85.65 \\
Without cross-event organization & 74.35 & 81.12 & 78.90 & 80.45 & 78.70 & 79.42 \\
\bottomrule
\end{tabular}
\end{table}

\subsection{LSGC: long-form narrative quality}
\label{app:lsgc-ablations}

The LSGC ablation uses the same five-starter protocol and model configuration
as the main evaluation. Each condition removes one NarraWorld component while
keeping the writing scaffold, memory cap, decoding settings, and accepted-history
update process fixed. The table reports the complete 16-criterion breakdown;
Overall is the equal-weight mean across criteria.

\begin{table*}[t]
\caption{LSGC ablations by criterion. Scores are direction-corrected macro
means over 15 samples (5 starters $\times$ 3 runs) on a 0--100 scale
($\uparrow$).}
\label{tab:lsgc-ablations}
\centering
\footnotesize
\setlength{\tabcolsep}{3pt}
\begin{tabular}{lrrrr}
\toprule
Criterion & Full NarraWorld & Without $B$ & Without $H$ & Without cross-event \\
\midrule
Believable Character Actions & 91.04 & 83.33 & \textbf{92.02} & 82.51 \\
Nuanced Characters & \textbf{91.22} & 81.93 & 90.09 & 81.16 \\
Pacing & \textbf{87.71} & 84.22 & 85.36 & 72.44 \\
World Building & 87.49 & 85.96 & 88.44 & 74.84 \\
Compelling Plot & \textbf{87.25} & 82.49 & 84.33 & 76.82 \\
Emotionally Engaging & \textbf{94.60} & 87.36 & 88.27 & 82.49 \\
Coherent & 92.02 & 88.36 & \textbf{92.11} & 75.05 \\
Weak Dialogue & \textbf{91.00} & 80.07 & 90.20 & 83.22 \\
Tell-Don't-Show & \textbf{86.87} & 85.49 & 86.07 & 79.56 \\
Unsurprising or Uncreative & \textbf{84.60} & 83.33 & 79.38 & 77.69 \\
Amateurish & \textbf{93.16} & 89.64 & 91.47 & 85.44 \\
Purple Prose & 89.67 & 87.80 & \textbf{90.69} & 84.64 \\
Forced Poetry or Metaphor & 88.56 & 88.18 & \textbf{90.29} & 87.89 \\
Unearned Transformations & \textbf{91.09} & 83.49 & 89.56 & 78.84 \\
Well-earned Lightness or Darkness & \textbf{92.42} & 88.71 & 86.07 & 82.22 \\
Faithful to Writing Prompt & 95.09 & \textbf{96.33} & 95.49 & 92.02 \\
\midrule
Overall & \textbf{90.24} & 86.04 & 88.74 & 81.05 \\
\bottomrule
\end{tabular}
\end{table*}

\begin{table}[t]
\caption{Run-to-run variability for LSGC ablations. Overall means are
direction-corrected scores; range and sample SD are computed over the three
independent runs after aggregating each run over five starters and 16 criteria.}
\label{tab:lsgc-ablation-variability}
\centering
\small
\setlength{\tabcolsep}{3pt}
\begin{tabular}{lrrr}
\toprule
Setting & Overall mean $\uparrow$ & Run range & Run SD \\
\midrule
Full NarraWorld & \textbf{90.24} & 2.59 & 1.35 \\
Without $B$ & 86.04 & 3.26 & 1.68 \\
Without $H$ & 88.74 & \textbf{1.53} & \textbf{0.80} \\
Without cross-event & 81.05 & 4.02 & 2.09 \\
\bottomrule
\end{tabular}
\end{table}

The ablation pattern is consistent across the aggregate and the detailed
criteria. Removing cross-event organization produces the largest Overall drop,
from $90.24$ to $81.05$ (a decrease of $9.19$ points), with especially large
losses in pacing, world building, and coherence. Removing character beliefs
reduces Overall to $86.04$ (a decrease of $4.20$ points) and is most visible
in the character-related criteria, while removing possible worlds has a smaller
aggregate effect ($88.74$) but changes several creativity and world-modeling
criteria. The full system therefore benefits from the complementary views:
cross-event structure supports long-range organization, character beliefs
preserve perspective, and possible worlds retain unresolved alternatives. The
run-level statistics also reveal a stability trade-off: removing $H$ produces
the smallest range and SD (1.53 and 0.80), whereas the full system is more
variable (2.59 and 1.35); removing cross-event organization is both the weakest
and the most variable ablation (4.02 and 2.09).

\subsection{LoCoMo: QA-time memory-view access}
\label{app:locomo-view-ablation}

LoCoMo uses a task-specific memory view because hypothetical-world records are
not necessarily admissible evidence for a factual answer. The default QA view
therefore excludes $H$, while the comparison view exposes it. The remaining
rows remove the same components used in the writing ablations. All conditions
use the shared 8k memory budget and the same category-aware scorer; the table
reports the complete subset breakdown.

\begin{table*}[t]
\caption{LoCoMo memory-view and component ablations at the shared 8k budget.
The default QA view excludes hypothetical-world records ($H$); the exposed-$H$
condition includes them. Overall Main aggregates the four reported question
subsets ($\uparrow$).}
\label{tab:locomo-full-ablation}
\centering
\small
\setlength{\tabcolsep}{4pt}
\begin{tabular*}{\textwidth}{@{\extracolsep{\fill}}lrrrrr@{}}
\toprule
Setting & \shortstack{Overall\\Main} & Multi-hop & Single-hop & Temporal & Open-domain \\
\midrule
NarraWorld (QA view: $H$ excluded) & \textbf{0.614} & \textbf{0.458} & \textbf{0.679} & \textbf{0.649} & \textbf{0.392} \\
NarraWorld (QA view: $H$ included) & 0.570 & 0.398 & 0.655 & 0.559 & 0.366 \\
Without character beliefs ($B$) & 0.586 & 0.415 & 0.650 & 0.635 & 0.370 \\
Without cross-event organization & 0.537 & 0.388 & 0.618 & 0.512 & 0.341 \\
\bottomrule
\end{tabular*}
\end{table*}

The QA-view comparison isolates access to $H$. Excluding $H$ improves Overall
Main by 4.4 points and improves every reported subset. This task dependence
complements the writing ablations, where $H$ remains available to preserve
unresolved alternatives during continuation.

\section{Human Evaluation and Judge Alignment}
\label{app:human-evaluation}

\subsection{Evaluation Protocol}

We paired human ratings with automatic scores on the same anonymized story
sample. For ConStory-Bench, we randomly sampled 12 prompts across the four
task types with equal allocation, comparing NarraWorld, Graphiti, and Full Context
on 36 stories. For LSGC, we evaluated NarraWorld, StructMem, and Full Context
on all five official starters, yielding 15 complete 14-chapter stories. This
human-evaluation sample is separate from the three-run aggregate used for the
main LSGC results.

Four native-English-speaking assessors with backgrounds in narrative analysis
evaluated anonymized texts without access to system identities or automatic
scores. Two assessors independently read each complete story. For ConStory,
they labeled the presence of each of the 19 error subtypes and supplied
textual evidence. A coordinator, also blind to system identities and automatic
results, adjudicated disagreements. For LSGC, readers evaluated complete stories
on five 0--100 dimensions: Character, Plot, World/Coherence, Style/Creativity,
and Prompt Adherence. The human rubric groups the 16 automatic criteria into
these five dimensions (Section~\ref{app:human-rubric}). Human Overall preserves
the relative weighting of the original 16-criterion rubric: the five dimension
scores receive weights of 3, 4, 2, 6, and 1, respectively, and the weighted sum
is divided by 16.

Across the 684 ConStory story--subtype pairs, both assessors marked seven
positive and 666 negative; five were positive only for assessor A and six
only for assessor B. Pooled agreement was 98.4\%, with Cohen's
$\kappa=0.55$. Positive agreement was 56.0\%
($2\times7/(2\times7+5+6)$), distinguishing agreement on errors from the
many shared negative labels. Adjudication yielded five positive labels.

\subsection{ConStory-Bench: Human Consistency Assessment}

Table~\ref{tab:constory-human} reports human and automatic CED on the same
12 prompts. Human assessment identified one positive error-subtype label for
each of NarraWorld and Full Context, compared with three for Graphiti. Human CED was
0.095, 0.289, and 0.096 for NarraWorld, Graphiti, and Full Context, respectively.
NarraWorld and Full Context therefore showed similar observed consistency, with
fewer confirmed error subtypes than Graphiti. The automatic judge flagged
more errors than human readers.

\begin{table*}[!htbp]
\caption{ConStory-Bench human evaluation and judge alignment on 12 prompts
(12 stories per method). $\Delta$ is Auto CED minus Human CED.
Detection metrics use adjudicated error-subtype presence as the reference.}
\label{tab:constory-human}
\centering
\scriptsize
\setlength{\tabcolsep}{4pt}
\begin{tabular}{lrrrrrrrrr}
\toprule
Method & Human CED $\downarrow$ & Auto CED $\downarrow$ & $\Delta$ & TP & FP & FN & Precision & Recall & F1 \\
\midrule
NarraWorld & 0.095 & 0.185 & +0.090 & 1 & 1 & 0 & 50.0\% & 100.0\% & 66.7\% \\
Graphiti & 0.289 & 0.385 & +0.096 & 2 & 2 & 1 & 50.0\% & 66.7\% & 57.1\% \\
Full Context & 0.096 & 0.200 & +0.104 & 1 & 1 & 0 & 50.0\% & 100.0\% & 66.7\% \\
\bottomrule
\end{tabular}
\end{table*}

Across the 36 stories, the automatic judge produced four false positives and
one false negative relative to the adjudicated human labels.

\subsection{LSGC: Human Quality and Score Alignment}

Human readers rated NarraWorld highest in long-form quality
(Table~\ref{tab:lsgc-human}), with an Overall score of 84.50, followed by
StructMem at 80.50 and Full Context at 80.06.
NarraWorld led Character, Plot, and Style/Creativity, whereas Full Context led 
World/Coherence and Prompt Adherence across the five official starters.

\begin{table*}[!htbp]
\caption{LSGC human scores across all five starters on a 0--100 scale
($\uparrow$). Auto Overall is shown alongside the human ratings for comparison.}
\label{tab:lsgc-human}
\centering
\footnotesize
\setlength{\tabcolsep}{2.5pt}
\begin{tabular}{lrrrrrrr}
\toprule
Method & Character & Plot & World/Coh. & Style/Creat. & Prompt & \shortstack{Human\\Overall} & \shortstack{Auto\\Overall} \\
\midrule
NarraWorld & \textbf{86.00} & \textbf{84.00} & 86.00 & \textbf{82.00} & 94.00 & \textbf{84.50} & \textbf{90.24} \\
StructMem & 84.00 & 81.00 & 82.00 & 76.00 & 92.00 & 80.50 & 86.92 \\
Full Context & 83.00 & 82.00 & \textbf{88.00} & 72.00 & \textbf{96.00} & 80.06 & 85.03 \\
\bottomrule
\end{tabular}
\end{table*}

Table~\ref{tab:lsgc-human-alignment} summarizes the alignment between the automatic 
judge's aggregated 5-dimension scores and the humans' direct 5-dimension assessments 
for the 15 complete stories. Overall correlation was moderate ($\rho=0.55$), with 
an average automatic score 5.71 points higher than the human score. Correlation
was lowest for Style/Creativity ($\rho=0.38$).

\begin{table}[!htbp]
\caption{LSGC human--judge alignment across 15 stories. Bias is the mean
automatic minus human score. Both MAE columns use the 0--100 scale.}
\label{tab:lsgc-human-alignment}
\centering
\small
\setlength{\tabcolsep}{6pt}
\begin{tabular}{lrrrr}
\toprule
Dimension & Spearman $\rho$ & MAE & Bias & \shortstack{Human--Human\\MAE} \\
\midrule
Character & 0.52 & 5.40 & +4.89 & 6.20 \\
Plot & 0.48 & 6.80 & +6.42 & 7.50 \\
World/Coherence & 0.61 & 5.10 & +4.67 & 5.50 \\
Style/Creativity & 0.38 & 7.60 & +7.17 & 9.00 \\
Prompt & 0.68 & 3.20 & +2.33 & 4.20 \\
Overall & 0.55 & 6.20 & +5.71 & 6.48 \\
\bottomrule
\end{tabular}
\end{table}

Human readers also differed on Plot and Style/Creativity, with human--human
MAE of 7.50 and 9.00. These differences reflect the subjective component of
literary assessment and provide context for the moderate judge alignment.

\subsection{Human Evaluation Rubric for Long-Form Narrative}
\label{app:human-rubric}

The human rubric groups the 16 Longform Writing Bench criteria
\citep{paech2025longform} into five dimensions. Assessors read each complete
story and assigned a score from 0 to 100 for each dimension using the following
guidelines.

\begin{swpmprompt}{Human Evaluator Guidelines: 5-Dimension Rubric}
\textbf{1. Character (0-100):} 
Evaluates the depth, consistency, and psychological realism of the cast.
\begin{itemize}
    \item \textit{High score criteria:} Characters have nuanced, distinct voices and 
    multi-dimensional personalities. Their actions, decisions, and emotional responses 
    are believable and strictly grounded in their established traits and past experiences.
    \item \textit{Penalties:} Cardboard or interchangeable characters; actions that break 
    character logic purely to advance the plot; sudden, unearned psychological transformations 
    or skill acquisitions that lack proper narrative buildup.
\end{itemize}

\textbf{2. Plot (0-100):} 
Evaluates the narrative arc, emotional engagement, and structural pacing.
\begin{itemize}
    \item \textit{High score criteria:} The story presents a compelling, emotionally 
    engaging trajectory with well-earned stakes. The pacing feels natural—slowing down 
    for character moments and accelerating for tension. Tonal shifts (lightness or darkness) 
    feel earned and impactful.
    \item \textit{Penalties:} Meandering, rushed, or sluggish pacing; disconnected episodes 
    lacking an overarching drive; emotional beats that feel manipulative, unearned, or hollow.
\end{itemize}

\textbf{3. World/Coherence (0-100):} 
Evaluates the solidity, logic, and internal consistency of the storyworld.
\begin{itemize}
    \item \textit{High score criteria:} The setting is vividly established and operates 
    under clear, consistent rules. The narrative flows logically from scene to scene, 
    tracking physical objects, locations, and time accurately across the entire text.
    \item \textit{Penalties:} Contradictory facts (e.g., dead characters reappearing, 
    locations shifting inexplicably); thin, generic world-building that lacks sensory 
    detail or internal logic; confusing spatial or temporal jumps.
\end{itemize}

\textbf{4. Style/Creativity (0-100):} 
Evaluates prose quality, narrative voice, and originality.
\begin{itemize}
    \item \textit{High score criteria:} The writing is engaging, professional, and 
    employs "show, don't tell" effectively. Dialogue is sharp and subtextual. 
    The narrative offers fresh, unsurprising twists or creative interpretations of tropes.
    \item \textit{Penalties:} "Amateurish" prose (stilted phrasing, repetitive sentence 
    structures); "purple prose" or overly forced, melodramatic metaphors; heavy exposition 
    dumps instead of organic action; generic, highly predictable, or cliché-ridden writing.
\end{itemize}

\textbf{5. Prompt Adherence (0-100):} 
Evaluates fidelity to the original writing premise and constraints.
\begin{itemize}
    \item \textit{High score criteria:} The story fully embraces and explores the core 
    concept, characters, and specific constraints dictated by the original prompt or starter.
    \item \textit{Penalties:} Ignoring key constraints; drifting entirely away from the 
    requested genre or premise; failing to utilize the provided opening effectively.
\end{itemize}
\end{swpmprompt}

\section{Limitations}
\label{app:limitations}

\paragraph{Modeling and construction cost.}
NarraWorld's richer structured modeling is more expensive than a compact
passage store, but its construction overhead remains comparable to other
structured or graph-based memory systems: baseline workloads span roughly 20
to 800 successful modeling calls per story, while NarraWorld averages about
200. This reflects a quality--cost trade-off: the current system prioritizes
explicit narrative structure, while reducing extraction and planning overhead
is an important direction for future work.

\paragraph{The contribution of possible worlds is conditional.}
The $H$ view is designed for long-form writing, where retaining unresolved
alternatives can support continuation quality, but hypothetical content is not
equally useful in general-purpose tasks such as factual QA. NarraWorld therefore
uses task-conditioned access to $H$; learning a more adaptive gate for this
writing-specific view is a promising direction for future work.

\section{Tracing Memory into Narrative: Qualitative Case Studies}
\label{app:qualitative-memory}

We examine how remembered details, relationships, and perspectives reappear
in ConStory-Bench and LSGC continuations from Mem0, Graphiti, StructMem, and
NarraWorld. The figures visualize selected textual memory records as condensed
structures; these diagrams were not writer inputs. Each method continues its
own generated history under a shared task and also receives recent text.
Selected for clear memory--narrative correspondences and contrasting outcomes,
these cases illustrate how memory is used within individual trajectories.
They do not estimate how often these patterns occur or isolate retrieval's
causal contribution. ConStory examples pair stored memory with continuations;
LSGC examples also examine the retrieved text presented to the writer.

\subsection{ConStory-Bench: Objects, Promises, and Reconciliation}
\label{app:qualitative-constory}

\begin{figure*}[!htbp]
\centering
\includegraphics[width=\textwidth]{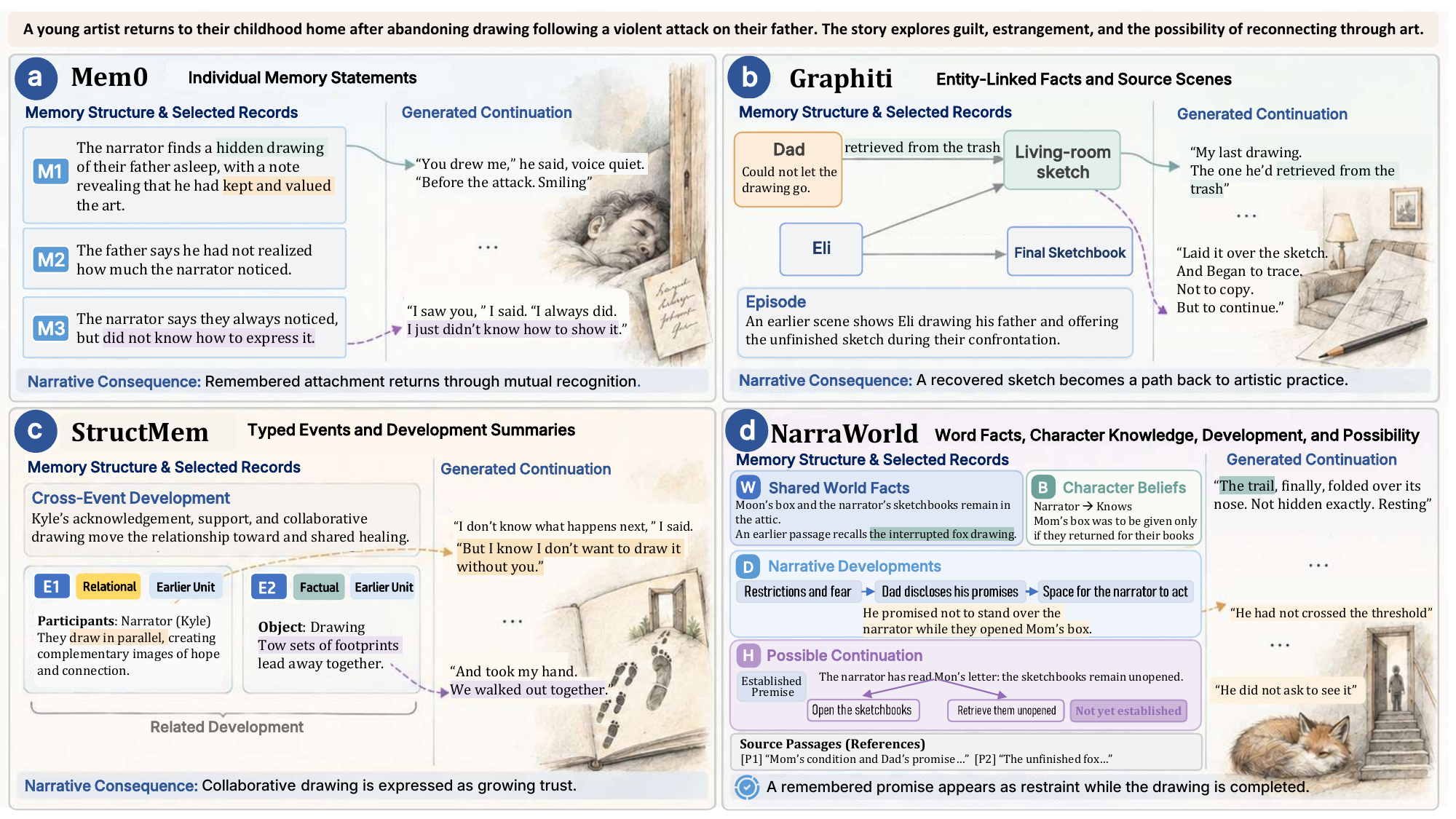}
\caption{Memory and continuation in ConStory-Bench prompt 128, unit 5.
Solid colored arrows connect recurring details; dashed arrows connect related
developments.}
\label{fig:constory-memory-narrative}
\end{figure*}

Figure~\ref{fig:constory-memory-narrative} shows prompt 128, continuation
unit 5. A young artist returns to their childhood home after abandoning
drawing following an attack on their father. Earlier drawings and family
promises acquire new significance as the story approaches reconciliation.

Mem0 retains the father's attachment to an old drawing and the narrator's
difficulty expressing affection; the continuation returns to these details
through mutual recognition. Graphiti links the father, a recovered sketch,
and an earlier confrontation. The sketch becomes an object the narrator can
resume, turning retrieval of a physical artifact into renewed artistic
practice. StructMem connects collaborative drawing to growing trust with Kyle;
the continuation expresses that relationship through the wish to keep drawing
together. These examples show distinct forms of continuity across statements,
entity relations, and developments.

NarraWorld connects the unfinished fox drawing, the narrator's knowledge of their
mother's condition for receiving her box, and the father's promise to give
them space. The continuation completes the fox while the father remains
outside the threshold and does not ask to see the drawing. A remembered
commitment thus appears as restraint in the scene, alongside a concrete
artistic callback. A remaining inconsistency is the sketchbook cover changing
from green to blue within NarraWorld's story. The case illustrates relational and
thematic continuity while leaving object-state tracking as a separate challenge.

\subsection{LSGC: Evidence, Perspective, and Reassessing Loyalty}
\label{app:qualitative-lsgc}

\begin{figure*}[!htbp]
\centering
\includegraphics[width=\textwidth]{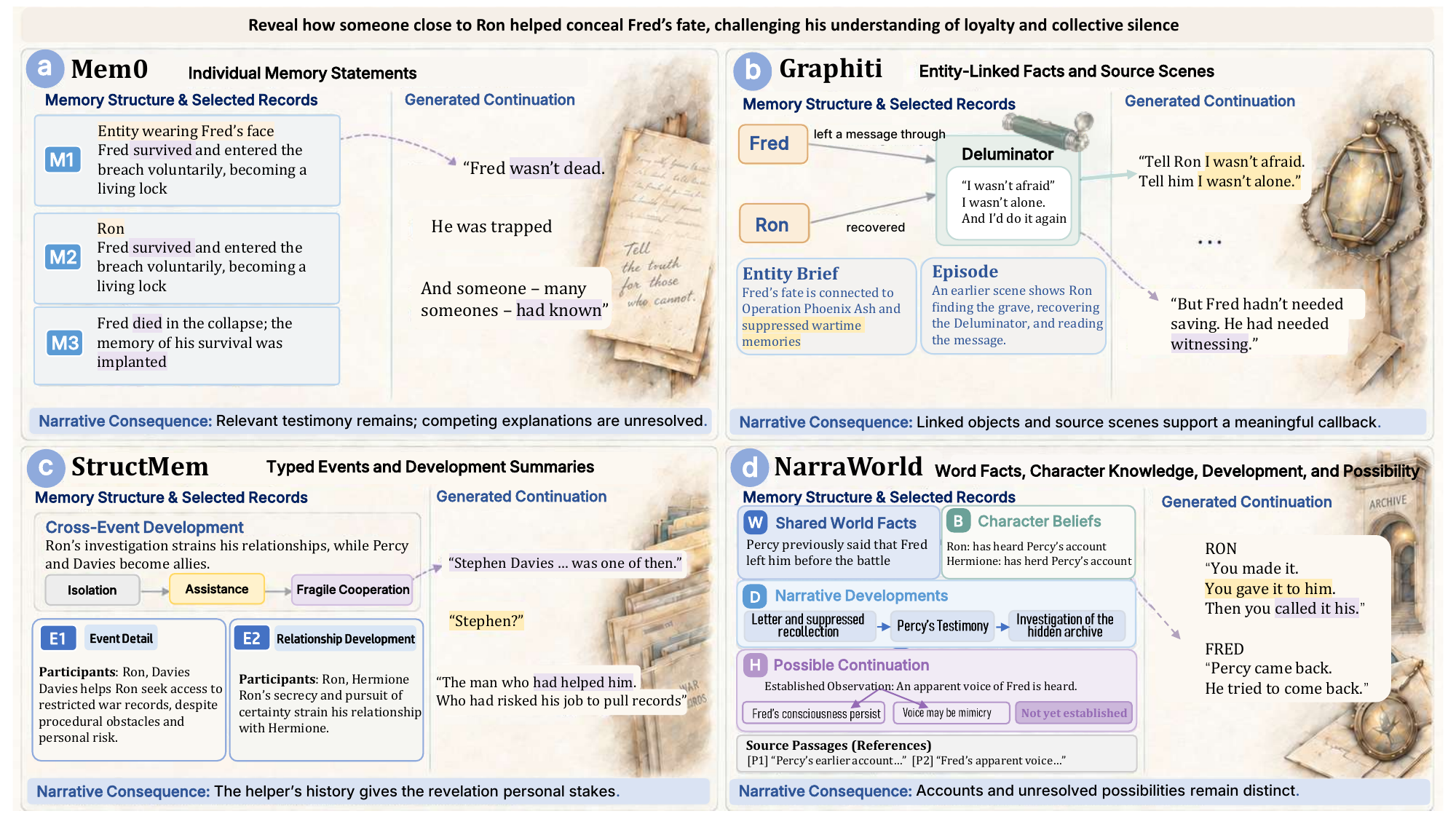}
\caption{Memory and revelation in LSGC starter-01, Chapter 10.
Solid arrows connect recurring content; dashed arrows connect related
developments and reinterpretations.}
\label{fig:memory-into-narrative}
\end{figure*}

We examine starter-01, Chapter 10, selected for its contrasting chapter-level
outcomes. The shared objective reveals that someone close to Ron helped
conceal Fred's fate, challenging his understanding of loyalty.
Figure~\ref{fig:memory-into-narrative} pairs each method's memory from the
preceding nine chapters with its continuation.

Mem0 retrieves attributed claims that Fred survived, that he died, and that
his apparent voice exploits Ron's grief, without explicit revision relations.
The continuation opens with Ron's conviction that Fred is trapped, but leaves
the competing accounts insufficiently reconciled. Graphiti exposes
entity-linked evidence and earlier scenes. Its continuation follows records
and testimony, returning to Fred's earlier message with a new emotional
meaning: Ron recognizes a need to bear witness. The object and its message
connect the investigation to a change in Ron's understanding.

StructMem preserves relationship changes across events. Davies's earlier
willingness to risk helping Ron gives the revelation of his complicity
personal stakes; Hermione's hurt over Ron's secrecy informs their renewed
cooperation. NarraWorld separately presents Percy's earlier account, identifies who
heard it, and tracks the developing investigation and unresolved interpretations
of Fred's apparent voice. Its continuation reframes Percy's concealment as a coerced
choice. Ron's response---``You made it. You gave it to him. Then you called
it his.''---shifts responsibility toward those who imposed that choice.
The revelation develops through a reassessment of prior testimony and
relationships.

Evidence updates remain a challenge: Graphiti's successive accounts of Fred's
fate are incompletely reconciled, while NarraWorld's continuation only partly
distinguishes evidence from Ron's interpretation
(Section~\ref{app:qualitative-limitations}).
Mem0 also exhibits an unexplained location transition, and StructMem reverses
a decision to stay behind without explanation. These observations distinguish
the availability of historical context from its use in a new scene.

\subsection{Further Examples of Constraints and Consequences}
\label{app:qualitative-additional}

The following cases extend the analysis to physical rules, responsibility,
evidence revision, and changing access to another character's inner state.
Each comparison follows the constraints established by its own story.

\paragraph{World rules constrain rescue actions.}
In ConStory prompt 1939, unit 6, NarraWorld's memory preserves two unusual safety
requirements: a containment indicator must remain red, and the capsule must
stay sealed. The continuation administers treatment through a sealed channel,
keeps the indicator red, and maintains separation while Mara reassures Nia
through the glass. Physical safety and emotional commitment remain compatible.
Graphiti's version connects a transmission network to an intervention through
the host, while StructMem develops cooperation with the quarantined girl.
Each trajectory follows its own established rescue mechanism.

\paragraph{Commitments become observable responsibility.}
In ConStory prompt 199, unit 8, memory records Cam changing a household reminder
from ``Kitchen reset. Do not make Nora ask.'' to ``Kitchen reset.'' It also
retains his promise to take responsibility for chores. The continuation makes
that responsibility explicit: Cam treats reminders as his own obligations,
while Eli acknowledges that listening should not depend on Nora remaining
calm. Later actions, including laundry and hanging a wet towel, carry the
conversation into daily behavior. Graphiti and StructMem also depict concrete
follow-through; NarraWorld's distinctive emphasis here is the allocation of
different responsibilities among three people.

\paragraph{New evidence changes the meaning of old clues.}
In ConStory prompt 461, unit 5, saved records distinguish Froshi's hopeful
interpretation of a bridle rein from Mara's caution. Witness accounts and a
recognized pendant subsequently change the interpretation of the rein, smoke,
and red wool: they guide people toward one another without establishing that
Froshi's sister remains alive. His grief and request for companionship persist
after that revision. Attributed interpretations in world records and their
connected developments support this revision.

\paragraph{Local success leaves consequences unresolved.}
In LSGC starter-02, Chapter 11, retrieved memory preserves the three tokens'
states and earlier conditions involving voluntary contributions and a witness.
The continuation uses the objects to establish a temporal anchor, but leaves
the warden undefeated and the house beginning to dissolve. This preserves the
requested crisis at the chapter boundary. Mem0 and Graphiti also depict the
initial confrontation, then advance toward reconciliation or departure.
The continuations differ in how far they advance beyond the chapter's crisis.

\paragraph{A relationship can continue after automatic access ends.}
In LSGC starter-04, Chapter 13, NarraWorld's memory connects coercive family
arrangements, warnings about free choice, and the protagonists' loss of privacy.
After the ritual, Pansy no longer automatically feels Theo's emotions and
expresses her choice through speech: ``This time, no magic carried the words.''
The continuation changes access to another mind while preserving the
relationship. Other methods develop stronger or transformed bonds, offering
different resolutions to the shared objective of attempting to break the bond.

\subsection{Qualitative Limitations}
\label{app:qualitative-limitations}

\paragraph{Representing uncertainty does not ensure scrutiny.}
The LSGC starter-01 case exposes a gap between retaining uncertainty and acting
on it during writing. Retrieved memory leaves open whether Fred's
voice reflects surviving consciousness or mimicry. A source passage also warns
that a reflector may show either truth or what its observer fears. Later,
images associated with Carrow and partly corroborating testimony rapidly
reshape Ron's judgment of Bill and Kingsley. The scene advances the emotional
conflict, but only partly distinguishes what each source supports and what
remains Ron's provisional interpretation. The scene illustrates the gap
between retaining uncertainty in memory and examining new evidence during
generation.

\paragraph{Remembered constraints still require execution.}
The sketchbook-color change in the ConStory case illustrates that thematic
continuity can coexist with inconsistent object state. Similarly, returning
relevant people and unresolved questions does not specify how a scene should
test competing explanations. Rapid revelation through dialogue can achieve a
chapter's emotional objective while leaving little space for investigation.
These observations motivate evaluating whether selected evidence becomes
appropriate action, belief revision, and state change, rather than measuring
retrieval alone. Targeted checks of consequential constraints are a direction
for future work.

Shared writing summaries should likewise preserve claim attribution and
uncertainty, since stronger summary wording can override distinctions retained
in memory.

\end{document}

%% file: math_commands.tex
\usepackage{amsmath,amsfonts,bm}

\def\eqref#1{equation~\ref{#1}}
\def\1{\bm{1}}

\DeclareMathAlphabet{\mathsfit}{\encodingdefault}{\sfdefault}{m}{sl}
\SetMathAlphabet{\mathsfit}{bold}{\encodingdefault}{\sfdefault}{bx}{n}

%% file: references.bib
@incollection{ryan2013possibleworlds,
  title     = {Possible Worlds},
  author    = {Ryan, Marie-Laure},
  editor    = {H{\"u}hn, Peter and others},
  booktitle = {The Living Handbook of Narratology},
  publisher = {Hamburg University},
  address   = {Hamburg},
  year      = {2013},
  url       = {https://www-archiv.fdm.uni-hamburg.de/lhn/node/54.html}
}

@inproceedings{piper2021narrative,
  title     = {Narrative Theory for Computational Narrative Understanding},
  author    = {Piper, Andrew and So, Richard Jean and Bamman, David},
  booktitle = {Proceedings of the 2021 Conference on Empirical Methods in Natural Language Processing},
  pages     = {298--311},
  year      = {2021},
  doi       = {10.18653/v1/2021.emnlp-main.26},
  url       = {https://aclanthology.org/2021.emnlp-main.26/}
}

@book{genette1983narrative,
  title     = {Narrative Discourse: An Essay in Method},
  author    = {Genette, G{\'e}rard},
  publisher = {Cornell University Press},
  year      = {1983}
}

@book{herman2009basic,
  title     = {Basic Elements of Narrative},
  author    = {Herman, David},
  publisher = {Wiley-Blackwell},
  year      = {2009}
}

@book{bal2009narratology,
  title     = {Narratology: Introduction to the Theory of Narrative},
  author    = {Bal, Mieke},
  translator = {Van Boheemen, Christine},
  publisher = {University of Toronto Press},
  year      = {2009}
}

@inproceedings{kearns2020narrativetime,
  title     = {Annotating and Quantifying Narrative Time Disruptions in Modernist and Hypertext Fiction},
  author    = {Kearns, Edward},
  booktitle = {Proceedings of the First Joint Workshop on Narrative Understanding, Storylines, and Events},
  pages     = {72--77},
  year      = {2020},
  address   = {Online},
  publisher = {Association for Computational Linguistics}
}

@inproceedings{bae2011focalization,
  title     = {Toward a Computational Model of Focalization in Narrative},
  author    = {Bae, Byung-Chull and Cheong, Yun-Gyung and Young, R. Michael},
  booktitle = {Proceedings of the 6th International Conference on the Foundations of Digital Games},
  pages     = {313--315},
  year      = {2011},
  doi       = {10.1145/2159365.2159423},
  url       = {https://doi.org/10.1145/2159365.2159423}
}

@inproceedings{teutenberg2015knowledge,
  title     = {Incorporating Global and Local Knowledge in Intentional Narrative Planning},
  author    = {Teutenberg, Jonathan and Porteous, Julie},
  booktitle = {Proceedings of the 14th International Conference on Autonomous Agents and Multiagent Systems},
  pages     = {1539--1546},
  year      = {2015},
  url       = {https://www.ifaamas.org/Proceedings/aamas2015/aamas/p1539.pdf}
}

@inproceedings{sanghrajka2022headspace,
  title     = {{HeadSpace}: Incorporating Action Failure and Character Beliefs into Narrative Planning},
  author    = {Sanghrajka, Rushit and Young, R. Michael and Thorne, Brandon},
  booktitle = {Proceedings of the AAAI Conference on Artificial Intelligence and Interactive Digital Entertainment},
  volume    = {18},
  pages     = {171--178},
  year      = {2022},
  doi       = {10.1609/aiide.v18i1.21961},
  url       = {https://doi.org/10.1609/aiide.v18i1.21961}
}

@inproceedings{han2022flashback,
  title     = {Go Back in Time: Generating Flashbacks in Stories with Event Temporal Prompts},
  author    = {Han, Rujun and Chen, Hong and Tian, Yufei and Peng, Nanyun},
  booktitle = {Proceedings of the 2022 Conference of the North American Chapter of the Association for Computational Linguistics: Human Language Technologies},
  pages     = {1450--1470},
  year      = {2022},
  doi       = {10.18653/v1/2022.naacl-main.104},
  url       = {https://aclanthology.org/2022.naacl-main.104/}
}

@misc{packer2023memgpt,
  title         = {{MemGPT}: Towards {LLM}s as Operating Systems},
  author        = {Packer, Charles and Wooders, Sarah and Lin, Kevin and Fang, Vivian and Patil, Shishir G. and Stoica, Ion and Gonzalez, Joseph E.},
  year          = {2023},
  eprint        = {2310.08560},
  archiveprefix = {arXiv},
  primaryclass  = {cs.AI},
  url           = {https://arxiv.org/abs/2310.08560}
}

@article{liu2023lostmiddle,
  title   = {Lost in the Middle: How Language Models Use Long Contexts},
  author  = {Liu, Nelson F. and Lin, Kevin and Hewitt, John and Paranjape, Ashwin and Bevilacqua, Michele and Petroni, Fabio and Liang, Percy},
  journal = {Transactions of the Association for Computational Linguistics},
  volume  = {12},
  pages   = {157--173},
  year    = {2024},
  doi     = {10.1162/tacl_a_00638},
  url     = {https://doi.org/10.1162/tacl_a_00638}
}

@inproceedings{ji2026reconstructed,
  title         = {Memory is Reconstructed, Not Retrieved: Graph Memory for {LLM} Agents},
  author        = {Ji, Shuo and Li, Yibo and Hooi, Bryan},
  booktitle     = {International Conference on Learning Representations},
  year          = {2026},
  url           = {https://openreview.net/pdf/061241ca5a30c124478594f14b2e19955327b0c0.pdf}
}

@misc{aluru2026magnet,
  title         = {From Personas to Plot: Character-Grounded Multi-Agent Story Generation for Long-Form Narratives},
  author        = {Aluru, Aayush and Ho, Chloe and Hammouri, Muhammad and Luo, Kerry and Malik, Myra and Lagasse, Ryan and Bahuguna, Arjun and Sharma, Vasu},
  year          = {2026},
  eprint        = {2607.00918},
  archiveprefix = {arXiv},
  primaryclass  = {cs.CL},
  url           = {https://arxiv.org/abs/2607.00918}
}

@inproceedings{wang2025dome,
  title     = {Generating Long-form Story Using Dynamic Hierarchical Outlining with Memory-Enhancement},
  author    = {Wang, Qianyue and Hu, Jinwu and Li, Zhengping and Wang, Yufeng and Li, Daiyuan and Hu, Yu and Tan, Mingkui},
  booktitle = {Proceedings of the 2025 Conference of the Nations of the Americas Chapter of the Association for Computational Linguistics},
  pages     = {1352--1391},
  year      = {2025},
  doi       = {10.18653/v1/2025.naacl-long.63},
  url       = {https://aclanthology.org/2025.naacl-long.63/}
}

@inproceedings{lyu2025facttrack,
  title     = {{FactTrack}: Time-Aware World State Tracking in Story Outlines},
  author    = {Lyu, Zhiheng and Yang, Kevin and Kong, Lingpeng and Klein, Dan},
  booktitle = {Proceedings of the 2025 Conference of the Nations of the Americas Chapter of the Association for Computational Linguistics: Human Language Technologies (Volume 1: Long Papers)},
  pages     = {2825--2848},
  year      = {2025},
  doi       = {10.18653/v1/2025.naacl-long.144},
  url       = {https://aclanthology.org/2025.naacl-long.144/}
}

@inproceedings{huot2025agentsroom,
  title     = {Agents' Room: Narrative Generation through Multi-step Collaboration},
  author    = {Huot, Fantine and Amplayo, Reinald Kim and Palomaki, Jennimaria and Jakobovits, Alice Shoshana and Clark, Elizabeth and Lapata, Mirella},
  booktitle = {International Conference on Learning Representations},
  pages     = {5150--5183},
  year      = {2025},
  url       = {https://proceedings.iclr.cc/paper_files/paper/2025/hash/0fbc8a83d93dd8021a4dd8d2d34138eb-Abstract-Conference.html}
}

@inproceedings{yang2023doc,
  title     = {{DOC}: Improving Long Story Coherence With Detailed Outline Control},
  author    = {Yang, Kevin and Klein, Dan and Peng, Nanyun and Tian, Yuandong},
  booktitle = {Proceedings of the 61st Annual Meeting of the Association for Computational Linguistics},
  pages     = {3378--3465},
  year      = {2023},
  doi       = {10.18653/v1/2023.acl-long.190},
  url       = {https://aclanthology.org/2023.acl-long.190/}
}

@inproceedings{yang2022re3,
  title     = {{Re3}: Generating Longer Stories With Recursive Reprompting and Revision},
  author    = {Yang, Kevin and Tian, Yuandong and Peng, Nanyun and Klein, Dan},
  booktitle = {Proceedings of the 2022 Conference on Empirical Methods in Natural Language Processing},
  pages     = {4393--4479},
  year      = {2022},
  doi       = {10.18653/v1/2022.emnlp-main.296},
  url       = {https://aclanthology.org/2022.emnlp-main.296/}
}

@inproceedings{mirowski2023dramatron,
  title     = {Co-Writing Screenplays and Theatre Scripts with Language Models: An Evaluation by Industry Professionals},
  author    = {Mirowski, Piotr and Mathewson, Kory W. and Pittman, Jaylen and Evans, Richard},
  booktitle = {Proceedings of the 2023 CHI Conference on Human Factors in Computing Systems},
  pages     = {1--34},
  year      = {2023},
  doi       = {10.1145/3544548.3581225},
  url       = {https://doi.org/10.1145/3544548.3581225}
}

@inproceedings{zhong2024memorybank,
  title     = {{MemoryBank}: Enhancing Large Language Models with Long-Term Memory},
  author    = {Zhong, Wanjun and Guo, Lianghong and Gao, Qiqi and Ye, He and Wang, Yanlin},
  booktitle = {Proceedings of the AAAI Conference on Artificial Intelligence},
  volume    = {38},
  pages     = {19724--19731},
  year      = {2024},
  doi       = {10.1609/aaai.v38i17.29946},
  url       = {https://ojs.aaai.org/index.php/AAAI/article/view/29946}
}

@inproceedings{chhikara2025mem0,
  title     = {{Mem0}: Building Production-Ready {AI} Agents with Scalable Long-Term Memory},
  author    = {Chhikara, Prateek and Khant, Dev and Aryan, Saket and Singh, Taranjeet and Yadav, Deshraj},
  booktitle = {ECAI 2025},
  pages     = {2993--3000},
  year      = {2025},
  doi       = {10.3233/FAIA251160},
  url       = {https://doi.org/10.3233/FAIA251160}
}

@misc{rasmussen2025zep,
  title         = {{Zep}: A Temporal Knowledge Graph Architecture for Agent Memory},
  author        = {Rasmussen, Preston and Paliychuk, Pavlo and Beauvais, Travis and Ryan, Jack and Chalef, Daniel},
  year          = {2025},
  eprint        = {2501.13956},
  archiveprefix = {arXiv},
  primaryclass  = {cs.AI},
  url           = {https://arxiv.org/abs/2501.13956}
}

@inproceedings{anokhin2025arigraph,
  title     = {AriGraph: Learning Knowledge Graph World Models with Episodic Memory for {LLM} Agents},
  author    = {Anokhin, Petr and Semenov, Nikita and Sorokin, Artyom and Evseev, Dmitry and Kravchenko, Andrey and Burtsev, Mikhail and Burnaev, Evgeny},
  booktitle = {Proceedings of the Thirty-Fourth International Joint Conference on Artificial Intelligence},
  pages     = {12--20},
  year      = {2025},
  doi       = {10.24963/ijcai.2025/2},
  url       = {https://www.ijcai.org/proceedings/2025/2}
}

@inproceedings{kang2025memoryos,
  title     = {Memory {OS} of {AI} Agent},
  author    = {Kang, Jiazheng and Ji, Mingming and Zhao, Zhe and Bai, Ting},
  booktitle = {Proceedings of the 2025 Conference on Empirical Methods in Natural Language Processing},
  pages     = {25961--25970},
  year      = {2025},
  doi       = {10.18653/v1/2025.emnlp-main.1318},
  url       = {https://aclanthology.org/2025.emnlp-main.1318/}
}

@inproceedings{latimer2026hindsight,
  title     = {Hindsight: Structured Agent Memory that Retains, Recalls, and Reflects},
  author    = {Latimer, Christopher and Boschi, Nicol\`o and Neeser, Andrew and Bartholomew, Chris and Srivastava, Gaurav and Wang, Xuan and Ramakrishnan, Naren},
  booktitle = {Proceedings of the 64th Annual Meeting of the Association for Computational Linguistics (Volume 3: System Demonstrations)},
  pages     = {275--285},
  year      = {2026},
  doi       = {10.18653/v1/2026.acl-demo.27},
  url       = {https://aclanthology.org/2026.acl-demo.27/}
}

@inproceedings{maharana2024locomo,
  title     = {Evaluating Very Long-Term Conversational Memory of {LLM} Agents},
  author    = {Maharana, Adyasha and Lee, Dong-Ho and Tulyakov, Sergey and Bansal, Mohit and Barbieri, Francesco and Fang, Yuwei},
  booktitle = {Proceedings of the 62nd Annual Meeting of the Association for Computational Linguistics},
  pages     = {13851--13870},
  year      = {2024},
  doi       = {10.18653/v1/2024.acl-long.747},
  url       = {https://aclanthology.org/2024.acl-long.747/}
}

@inproceedings{park2023generative,
  title     = {Generative Agents: Interactive Simulacra of Human Behavior},
  author    = {Park, Joon Sung and O'Brien, Joseph C. and Cai, Carrie J. and Morris, Meredith Ringel and Liang, Percy and Bernstein, Michael S.},
  booktitle = {Proceedings of the 36th Annual ACM Symposium on User Interface Software and Technology},
  year      = {2023},
  doi       = {10.1145/3586183.3606763},
  url       = {https://doi.org/10.1145/3586183.3606763}
}

@inproceedings{xu2026structmem,
  title     = {{StructMem}: Structured Memory for Long-Horizon Behavior in {LLM}s},
  author    = {Xu, Buqiang and Chen, Yijun and Fang, Jizhan and Zhong, Ruobin and Yao, Yunzhi and Zhu, Yuqi and Du, Lun and Deng, Shumin},
  booktitle = {Proceedings of the 64th Annual Meeting of the Association for Computational Linguistics (Volume 2: Short Papers)},
  pages     = {122--146},
  year      = {2026},
  doi       = {10.18653/v1/2026.acl-short.12},
  url       = {https://aclanthology.org/2026.acl-short.12/}
}

@misc{paech2025longform,
  author        = {Paech, S. J.},
  title         = {Longform Creative Writing Benchmark},
  year          = {2025},
  url           = {https://github.com/EQ-bench/longform-writing-bench},
  note          = {GitHub repository}
}

@inproceedings{li2026constory,
  title     = {Lost in Stories: Consistency Bugs in Long Story Generation by {LLM}s},
  author    = {Li, Junjie and Guo, Xinrui and Wu, Yuhao and Lee, Roy Ka-Wei and Li, Hongzhi and Xie, Yutao},
  booktitle = {Findings of the Association for Computational Linguistics: ACL 2026},
  pages     = {8400--8428},
  year      = {2026},
  doi       = {10.18653/v1/2026.findings-acl.410},
  url       = {https://aclanthology.org/2026.findings-acl.410/}
}

@inproceedings{xiang2025rmtbench,
  title         = {{RMTBench}: Benchmarking {LLM}s Through Multi-Turn User-Centric Role-Playing},
  author        = {Xiang, Hao and Tang, Tianyi and Su, Yang and Yu, Bowen and Yang, An and Huang, Fei and Zhang, Yichang and Lu, Yaojie and Lin, Hongyu and Han, Xianpei and Zhou, Jingren and Lin, Junyang and Sun, Le},
  booktitle     = {Findings of the Association for Computational Linguistics: EMNLP 2025},
  pages         = {13555--13571},
  year          = {2025},
  doi           = {10.18653/v1/2025.findings-emnlp.730},
  url           = {https://aclanthology.org/2025.findings-emnlp.730/}
}

@inproceedings{migal2024lsgc,
  title     = {Overview of Long Story Generation Challenge ({LSGC}) at {INLG} 2024},
  author    = {Migal, Aleksandr and Seredina, Daria and Telnina, Ludmila and Nazarov, Nikita and Kolmogorova, Anastasia and Mikhaylovskiy, Nikolay},
  booktitle = {Proceedings of the 17th International Natural Language Generation Conference: Generation Challenges},
  pages     = {47--53},
  year      = {2024},
  doi       = {10.18653/v1/2024.inlg-genchal.4},
  url       = {https://aclanthology.org/2024.inlg-genchal.4/}
}
